\documentclass[sn-mathphys-num]{sn-jnl}% Math and Physical Sciences Numbered Reference Style 
\usepackage{graphicx}%
\usepackage{multirow}%
\usepackage{amsmath,amssymb,amsfonts}%
\usepackage{amsthm}%
\usepackage{mathrsfs}%
\usepackage[title]{appendix}%
\usepackage{xcolor}%
\usepackage{textcomp}%
\usepackage{booktabs}%
\usepackage{algorithm}%
\usepackage{algorithmicx}%
\usepackage{algpseudocode}%
\usepackage{listings}%
\usepackage{longtable}
\usepackage{rotating}

\usepackage{mathtools}          % Extends amsmath, providing common math tools
\usepackage{mathrsfs}           % Enables \mathscr, which can work in cases that \mathcal does not
\mathtoolsset{showonlyrefs}     % Only number equations that are referenced (optional)
\usepackage{subcaption}         % Allows for the use of subfigures and subcaptions
\usepackage[space]{grffile}     % For spaces in image names
\usepackage{url}                % For displaying urls
\usepackage{enumitem}
\usepackage{booktabs}       % professional-quality tables
\usepackage{array} % Add this line to your preamble
\usepackage{placeins}
\definecolor{commentcolor}{RGB}{0, 102, 204} % A medium blue color
\usepackage{verbatim}
\usepackage[export]{adjustbox}
\newcommand{\review}[1]{#1}

\begin{document}

\title[Article Title]{The impact of intrinsic rewards on exploration in Reinforcement Learning}

\author*{\fnm{Aya} \sur{Kayal}}\email{aya.kayal.21@ucl.ac.uk}

\author{\fnm{Eduardo} \sur{Pignatelli}}\email{e.pignatelli@ucl.ac.uk}

\author{\fnm{Laura} \sur{Toni}}\email{l.toni@ucl.ac.uk}

\affil{\orgdiv{Department of Electronic \& Electrical Engineering}, \orgname{University College London}}


\abstract{One of the open challenges in Reinforcement Learning (RL) is the hard exploration problem in sparse reward environments. Various types of intrinsic rewards have been proposed to address this challenge by pushing towards diversity. This diversity might be imposed at different levels, favouring the agent to explore different states, policies or behaviours (State, Policy and Skill level diversity, respectively). However, the impact of diversity on the agent's behaviour remains unclear.
In this work, we aim to fill this gap by studying the effect of different levels of diversity imposed by intrinsic rewards on the exploration patterns of RL agents. We select four intrinsic rewards (State Count, Intrinsic Curiosity Module (ICM), Maximum Entropy, and Diversity is all you need (DIAYN)), each pushing for a different diversity level. We conduct an empirical study on MiniGrid environment to compare their impact on exploration considering various metrics related to the agent's exploration, namely: episodic return, observation coverage, agent's position coverage, policy entropy, and timeframes to reach the sparse reward. The main outcome of the study is that State Count leads to the best exploration performance in the case of low-dimensional observations. However, in the case of RGB observations, the performance of State Count is highly degraded mostly due to representation learning challenges. Conversely, Maximum Entropy is less impacted, resulting in a more robust exploration, despite being not always optimal. % Conversely,  in the case of high-dimensional observations, Policy level diversity  is preferable due to its robustness to representation learning challenges.
%Lastly, our empirical study showed that learning diverse skills with DIAYN, often associated with enhanced robustness and generalisation, does not aid  exploration in MiniGrid environments.} The reason is that first it is hard to learn the skill space, second exploring in the skill space focus on distinguishing between different behaviours rather than visiting states as uniformly as possible.}
Lastly, our empirical study revealed that learning diverse skills with DIAYN, often linked to improved robustness and generalisation, does not promote exploration in MiniGrid environments. This is because: \emph{i)} learning the skill space itself can be challenging, and \emph{ii)} exploration within the skill space prioritises differentiating between behaviours rather than achieving uniform state visitation.}
\keywords{Reinforcement Learning, Intrinsic Motivation, Exploration, Diversity}

\maketitle

\section{Introduction}
\label{sec:Intro}
% The exploration problem + intrinsic rewards
%\review{RL has made significant progress in a variety of domains. For instance, RL studies have explored the adaptation of learning strategies to specific environmental or task contexts, with applications such as course recommendation, traffic recovery, and re-identification \citep{lin2022context,kiani2023novel,thakoor2013context}. While these studies highlight the recent success of RL and its adaptability, the field still suffers from a major hurdle: the sparsity of rewards \citep{ladosz2022exploration,abel2021bad}.}
The sparsity of rewards is a major hurdle for RL algorithms  \citep{ladosz2022exploration,abel2021bad}. 
With infrequent feedback, the probability of the agent randomly discovering a rewarding sequence of actions becomes low. Therefore, a large number of samples is needed to explore and stumble into a successful sequence of actions leading to the desired outcome \citep{gou2019dqn}. This is known as the hard exploration problem \citep{ladosz2022exploration}. Classical exploration strategies, e.g., epsilon-greedy and %Thompson sampling \citep{thompson1933likelihood},
Boltzmann distribution \citep{wiering1999explorations}  fail to explore the environment efficiently enough to find the optimal solution when the feedback is sparse \citep{franccois2018introduction}. Among the possible solutions to address this limitation \citep{ladosz2022exploration,amin2021survey,aubret2019survey}, intrinsic rewards \citep{singh2010intrinsically,chentanez2004intrinsically} have been proposed. They are a part of the larger notion of intrinsic motivation defined by \citet{ryan2000self} as the tendency to ``\textit{seek out novelty and challenges, to extend and exercise one’s capacity, to explore, and to learn}''.
% Taxonomy of intrinsic rewards 
Intrinsic rewards are often categorised in the literature into: \textit{knowledge-based} and \textit{competence-based} \citep{oudeyer2007intrinsic,colas2022autotelic,aubret2019survey,siddique2017review}. The first category encourages the agent to gain new knowledge about the environment. It compares the agent’s experiences to its existing knowledge, and rewards the agent for encountering unexpected situations.  %The first category pushes the agent to acquire new knowledge about the environment by comparing the situations experienced by the agent with the agent's predictions.
This includes methods that reward novelty in states or state transitions \citep{bellemare2016unifying,tang2017exploration,burda2018exploration, badia2020never}, the prediction error \citep{pathak2017curiosity} or the information gain \citep{houthooft2016vime}. The second category, also called ``\textit{skill learning}'' in \citet{aubret2019survey,aubret2023information}, rewards the agent for learning a diverse repertoire of skills in an unsupervised way. It mainly includes goal-conditioned RL approaches, which generate and achieve their own goals to explore the environment  \citep{eysenbach2018diversity,cohen2019maximum,gregor2016variational}. %\citep{aubret2019survey} subdivided this category into building the goal space from the state space ot maximizing mutual information between a goal and its associated trajectory.%
In \citet{colas2022autotelic}, a detailed survey on goal-conditioned RL is presented, highlighting the different types of goal representations and goal-sampling strategies. 

\review{This categorisation uncovers a potential link between diversity and exploration, where intrinsic rewards promote diverse agent behaviours to efficiently explore the environment. While diversity is acknowledged as crucial in RL, it has mainly been explored in relation to robustness, generalisation, hierarchical learning or generation tasks
~\citep{bettini2024controlling,osa2024discovering,kumar2020one,zahavy2022discovering,grillotti2024quality,mckee2022quantifying,chen2024dgpo, cideron2024diversity}. However, its role in driving effective exploration remains underexplored and not empirically validated yet. %Given that exploration is not mutually exclusive to generalisation, as shown in~\citep{jiang2024importance,korkmaz2024survey}, diversity might be
%a way to improve exploration beyond its other benefits. %Most studies on intrinsic rewards for exploration focus on either state novelty-driven (knowledge-based) or skill-driven (competence-based) approaches, yet do not compare these methods within a unified framework or examine how different levels of diversity influence exploration. Although prior surveys have suggested a connection between diverse skill learning and exploration, this connection has yet to be empirically validated. %Moreover, the empirical connection between diverse skill learning and exploration, as suggested by theoretical surveys, has not been thoroughly investigated.
%In this work, we aim to take the first step into advancing the understanding whether mechanisms that favor diversity through skill discovery can also drive effective exploration.
%In this work, we aim to go deeper in this analysis, and make the first dent into the matter of understanding if mechanisms that favor diversity through skill discovery can
%also push for good exploration.
%To fill this gap, we introduce a rigorous methodology to empirically compare knowledge-based and competence-based intrinsic rewards which was never done in the existing work, with a particular focus on how different levels of diversity affect exploration. 
In this work, we take an initial step toward understanding whether mechanisms that encourage diversity through skill discovery can also drive more effective exploration. To address this gap, we propose a rigorous methodology to empirically compare knowledge-based and competence-based intrinsic rewards, which has not been thoroughly investigated in prior research. Our work focuses on examining how different levels of diversity impact exploration, driven by the need to address the following open questions: \emph{i)} What is, in practice, an effective exploration in environments with low- and high-dimensional state spaces? 
 \emph{ii)} How does the level of diversity imposed by intrinsic rewards affect exploration performance across different scenarios?
 \emph{iii)} Does behavioural diversity through skill discovery, known to help robustness and fast adaptation \citep{kumar2020one,zahavy2022discovering}, also helps exploration? }
 
\review{\noindent Our contributions are twofold:
\begin{enumerate}
    \item We design an empirical study that categorises intrinsic rewards based on their level of diversity and assesses how these categories impact exploration using various metrics in a controlled setting. 
    \item We provide empirical insights on the role of diversity in exploration, offering practical guidance on how intrinsic rewards can be leveraged for environments with varying exploration challenges.
\end{enumerate}
Specifically, we conduct a comprehensive empirical study that builds on the existing categorisations of intrinsic rewards in the literature, further subdividing them into different ``diversity levels'' (State, State + Dynamics, Policy, and Skill levels of diversity). We select representative intrinsic reward methods from each level and evaluate their performance on  MiniGrid \citep{MinigridMiniworld23}, with both grid encodings and RGB observations. This setup enables us to investigate how different levels of diversity affect the agent behaviour in environments where exploration is crucial for task completion. Our study provides both qualitative and quantitative analyses of exploration across various metrics, including return, coverage, entropy, reward findings, and state visitation maps. To the best of our knowledge, this is the first study to systematically evaluate the impact of different diversity levels on exploration using a unified evaluation framework and incorporating multiple metrics, offering novel insights into how diversity influences exploration and performance in RL tasks.}
\section{Related Works}

\review{While numerous intrinsic reward formulations have been proposed to address complex sparse-reward tasks, a comprehensive understanding of their comparative advantages and challenges remains elusive, leaving this an open question in the field. Here, we review previous works that have attempted to categorise or empirically compare intrinsic rewards.  Table \ref{related_works_table} provides an overview of these studies, highlighting the pros and cons of each approach.}
\review{Existing surveys \citep{aubret2023information,ladosz2022exploration,aubret2019survey,amin2021survey,colas2022autotelic,hao2023exploration,siddique2017review} offer slightly different taxonomies of intrinsic rewards, often using varied terminology. However, most include two broad categories: one focused on increasing knowledge about the environment (e.g., prediction error, information gain, learning progress, and state novelty), and another focused on learning diverse skills. Yet, these surveys lack empirical validation and none of them explore the different levels of diversity that these intrinsic rewards can introduce within each category. In this work, we build on the categorisation proposed by \citep{colas2022autotelic}, which clearly distinguishes between knowledge-based and competence-based intrinsic rewards, and we further subdivide them into different diversity levels (state/dynamics/policy/skill).}  %However, none of the surveys gave insights on the different diversity levels that can be imposed by these intrinsic rewards, and what is the impact of such diversity on the exploration performance.%

We are now interested in the works provided in the literature aimed at benchmarking different
intrinsic rewards.
A few studies have compared methods within the knowledge-based category. For instance, \citet{andres2022evaluation} compared State Count \citep{strehl2008analysis}, Random Network Distillation (RND) \citep{burda2018exploration}, Intrinsic Curiosity Module (ICM) \citep{pathak2017curiosity}, Reward Impact Driven Exploration (RIDE) \citep{raileanu2020ride} on MiniGrid environment. The study aimed to evaluate the impact that weighting intrinsic rewards has on performance, as well as the effect of using different neural network architectures. Another study by \citet{taiga2021bonus} evaluated pseudo-counts \citep{bellemare2016unifying}, RND, ICM and Noisy Networks \citep{fortunato2017noisy} within the Arcade Learning Environment (ALE)~\citep{bellemare2013arcade}, and suggested that none of these methods outperform the epsilon-greedy exploration. 
\review{A more recent work by \citet{yuan2024rlexplore} introduced RLeXplore, a comprehensive plug-and-play framework that implements ICM~\citep{pathak2017curiosity}, RND~\citep{burda2018exploration}, Disagreement~\citep{pathak2019self}, Never Give Up (NGU)~\citep{badia2020never}, PseudoCounts~\citep{bellemare2016unifying}, RIDE~\citep{raileanu2020ride}, Random Encoders for Efficient Exploration (RE3)~\citep{seo2021state}, and Exploration via Elliptical Episodic Bonuses (E3B)~\citep{henaff2022exploration}. Their framework addressed critical design, implementation, and optimisation issues related to intrinsic rewards, including reward and observation normalisation, co-learning dynamics of policies and representations, weight initialisation, and the combined optimisation of intrinsic and extrinsic rewards.}
The study most similar to ours is by \citet{laskin2021urlb} which evaluated intrinsic rewards across knowledge, competence and data-based categories on the DeepMind Control Suite. However, their primary objective was to assess the generalisation of unsupervised RL algorithms by measuring how quickly they adapted to diverse downstream tasks. To achieve this, they used a reward-free pretraining phase followed by supervised finetuning. In contrast, our study focuses on the standard RL setting, where both intrinsic and extrinsic rewards are optimised simultaneously (except for skill-based learning). Instead of concentrating on adaptation, we address the exploration challenge, evaluating intrinsic rewards from a diversity perspective and employing various metrics to measure exploration quality. 
%It's important to note that generalisation and exploration are not mutualy exclusive and 
%However, the authors' primary objective was to evaluate the generalisation of unsupervised RL algorithms by measuring how quickly they adapt to diverse downstream tasks. Consequently, they employed a reward-free pretraining setting followed by supervised finetuning for downstream tasks. In contrast, our focus is on the RL setting, where both intrinsic and extrinsic rewards are optimised simultaneously (except for skill-based learning). Instead of addressing the adaptation efficiency, we tackle the exploration challenge by evaluating intrinsic rewards from a diversity perspective, employing various metrics to measure the quality of exploration.
 
% However, to study the unsupervised algorithms' generalisation capabilities, authors considered the reward-free pretraining setting followed by supervised finetuning to downstream tasks. In contrast, we consider the supervised RL setting, where both intrinsic and extrinsic rewards are optimised simultaneously, for all intrinsic reward methods except skill-based learning and we focus on addressing the exploration challenge from a diversity perspective.
Other works have examined a different taxonomy of intrinsic rewards: global vs. episodic bonuses. Global bonuses are calculated using the entire training experience, while episodic bonuses are calculated using only the experience from the current episode. The work by \citet{wang2022revisiting} found that episodic bonuses are more crucial than global bonuses to improve exploration in procedurally generated environments such as MiniGrid. A later study by \citet{henaff2023study} found that episodic bonuses tend to yield better results in situations where there is minimal shared structure across various contexts in MiniHack \citep{samvelyan2021minihack}, while global bonuses tend to be effective in cases where there is a greater degree of shared structure.

\review{Additionally, some works aimed to unify different intrinsic reward formulations under a general framework. For instance,  \citet{lin2024mimex} proposed a unified framework for intrinsic rewards, showing that existing methods can be viewed as special cases of conditional prediction with different mask distributions. Building on this, they introduced a novel trajectory-level exploration intrinsic reward, which extends beyond the typical one-step future prediction to capture transition dynamics across longer time horizons.  In a related line of work, \citet{zahavy2021reward} reformulated the convex MDP problem as a convex-concave game between an agent and an adversarial player generating costs (negative rewards). They unified a broad range of RL algorithms, including methods for unsupervised skill discovery, by interpreting them as instances of this generalised game-theoretic framework.} %This connects to the diversity-exploration open question, highlighting the fact that sequence-level intrinsic rewards which encourages agent to search for new trajectories, may lead to
%more complex exploration behavior and improve exploration efficiency, compared to one-step transitions.}
%Besides empirical studies, \citep{lin2024mimex} proposed a general framework for deriving intrinsic rewards, leads to a unified view on existing intrinsic reward approaches:
%they are special cases of conditional prediction, where the estimation of novelty can
%be seen as pseudo-likelihood estimation with different mask distributions. can be viewed as modeling different
%conditional prediction problems, or masked prediction problems with different mask distributions. Under this view, we can unify existing intrinsic reward methods into
%one framework, where only the underlying conditional prediction problem differs. MIMEx allows for flexible trajectory-level
%exploration via generalised conditional prediction.Existing approaches framed as
%conditional prediction often consider one-step future prediction problems. ; by setting up conditional prediction
%problems on trajectories, we can obtain intrinsic rewards that consider transition dynamics across
%longer time horizons and extract richer exploration signals. This encourages the agent to explore trajectory sequences that have been less frequently encountered}

Despite these significant advances in categorising, evaluating and interpreting intrinsic rewards in RL, a critical gap remains: the impact of diversity in intrinsic rewards on the exploration performance has not been thoroughly examined. Specifically, it is unclear how the exploration performance of competence-based methods, which encourage diverse behaviours, compares to knowledge-based methods that promote diverse states. In this study, we provide an initial empirical investigation into the impact of different levels of diversity on exploration across several MiniGrid environments, \review{serving as a preliminary effort to understand the complex relation between diversity and exploration in RL.}

\begin{sidewaystable} % Use sidewaystable for single-column width, sidewaystable* for double-column width
\centering

\begin{longtable}{|p{2cm}|p{3cm}|p{7cm}|p{5cm}|}
\caption{Comparison of existing works on intrinsic reward categorisations and empirical studies outlining the pros and cons.} 
\label{related_works_table} \\
\hline
\textbf{Study} & \textbf{Paper Category} & \textbf{Pros} & \textbf{Cons/Limitations} \\
\hline
\endfirsthead

\hline
\textbf{Study} & \textbf{Paper Category} & \textbf{Pros} & \textbf{Cons/Limitations} \\
\hline
\endhead

\citet{ladosz2022exploration} & Categorisation of intrinsic rewards & Categorised intrinsic rewards into reward novel states and reward diverse behaviours&  Lack of empirical testing in a common framework. \\
\hline
\citet{aubret2019survey} & Categorisation of intrinsic rewards & Categorised intrinsic rewards into knowledge acquisition and skill learning &  Lack of empirical testing in a common framework. \\
\hline
\citet{aubret2023information} & Categorisation of intrinsic rewards  & Categorised intrinsic rewards into surprise, novelty and skill learning&  Lack of empirical testing in a common framework. \\
\hline
\citet{amin2021survey} & Categorisation of intrinsic rewards  & Categorised intrinsic rewards into blind, uncertainty, space coverage and self-generated goals &  Lack of empirical testing in a common framework. \\
\hline
\citet{colas2022autotelic} & Categorisation of intrinsic rewards  & Categorised intrinsic rewards into knowledge and competence-based (focused on goal-conditioned RL) &  Lack of empirical testing in a common framework. \\
\hline
\citet{hao2023exploration} & Categorisation of intrinsic rewards  & Categorised intrinsic rewards into prediction-error, novelty and information gain &  Lack of empirical testing in a common framework. \\
\hline
\citet{siddique2017review} & Categorisation of intrinsic rewards  & Categorised intrinsic rewards into knowledge-based and competence-based &  Lack of empirical testing in a common framework. \\
\hline
\citet{andres2022evaluation} & Empirical study on intrinsic rewards& Evaluated performance of knowledge-based intrinsic rewards  (State Count, RND, ICM, RIDE) on MiniGrid. Analysed different weighting methods for intrinsic rewards and different neural network architectures. & Did not include any skill-learning methods from the competence-based category. Focused on return performance without directly studying exploration.  \\
\hline
\citet{taiga2021bonus} & Empirical study on intrinsic rewards&  Compared the performance of knowledge-based intrinsic rewards (Pseudo-counts, RND, ICM, Noisy Networks)  on ALE environment. & Did not include any skill learning/goal-conditioned methods. Evaluated performance solely based on return with no specific focus on the exploration behaviour. \\
\hline
\citet{yuan2024rlexplore} & Empirical study on intrinsic rewards&  Compared the performance of knowledge-based intrinsic rewards (ICM, RND, Disagreement, NGU, PseudoCounts, RIDE, RE3, and E3B). Addressed key design and optimisation details of intrinsic rewards to establish standardised implementations.& Did not include skill learning/goal-conditioned methods. Did not provide any link between diversity and exploration. \\
\hline
\citet{laskin2021urlb} & Categorisation and empirical study on intrinsic rewards & Tested methods from knowledge-based (ICM, Disagreement, RND), competence-based (DIAYN, SMM, APS) and data-based (APT, ProtoRL) categories on continuous control tasks. Evaluated their generalisation capabilities in an unsupervised pretraining followed by supervised finetuning framework. & Did not consider the joint optimisation of intrinsic and extrinsic rewards. Focused on generalisation and fast adaptation rather than studying the impact of diversity on exploration. \\
\hline
\citet{wang2022revisiting} & Categorisation and empirical study on intrinsic rewards & Divided intrinsic rewards between lifelong (global) and episodic bonuses. Tested different combinations of global and episodic bonuses on MiniGrid, in sparse reward and pure exploration settings. Analysed why lifelong intrinsic reward does not contribute much in improving exploration. & Focused on global vs episodic perspective, not on diversity and its impact on exploration. \\
\hline
\citet{henaff2023study} & Categorisation and empirical study of intrinsic rewards & Studied the advantages and disadvantages of global and episodic intrinsic rewards for exploration in contextual MDPs.  & Interpreted intrinsic rewards from the global/episodic perspective, but not from a diversity perspective. \\
\hline
\citet{lin2024mimex} & General framework unifying intrinsic rewards & Interpreted intrinsic rewards as special cases of conditional prediction with different mask distributions & Lack of a comparative empirical study \\
\hline
\citet{zahavy2021reward} & General framework unifying intrinsic rewards & Reformulated the convex MDP problem as a convex-concave game and interpreted several RL algorithms (including skill-based intrinsic rewards) as instances of it & Lack of a comparative empirical study \\
\hline
\end{longtable}
\end{sidewaystable}

%\textcolor{red}{To the best of our knowledge, none of these empirical studies compared the exploration performance of competence-based methods that push towards diverse behaviours with the knowledge-based ones that push towards diverse states, in a supervised RL setting}. Here, we provide an initial empirical study aimed at understanding the impact of different levels of diversity on exploration in several MiniGrid environments.
\section{Methodology}
In the following, we sub-classify the knowledge and competence-based intrinsic reward methods according to the level of diversity they impose on the agent's exploration (Section~\ref{subsection:diversity_levels}). Then, we select four intrinsic rewards, one for each level (Section~\ref{intrinsic rewards}), and we test them empirically on MiniGrid environment, explained and motivated in Section~\ref{subsection:Environment}. \review{Section~\ref{subsection:Experimental_Protocol} outlines the experimental protocol used in the study, while  Section~\ref{architecture} details the model architecture. Finally, Section~\ref{metrics} introduces the evaluation metrics.}

\subsection{Taxonomy of diversity levels imposed by intrinsic reward} \label{subsection:diversity_levels}
We systematise the type of diversity imposed by intrinsic rewards into four levels: \textbf{State level diversity} encourages exploration of unseen states, pushing the agent towards areas where its knowledge is most limited. \textbf{State + Dynamics level diversity} also focuses on diverse states, but additionally considers the novelty of the dynamics between those states for a more comprehensive exploration. \textbf{Policy level diversity} explores the impact of different actions from given states, while \textbf{Skill level diversity} explores the effectiveness of diverse skills (policy-goal association) in achieving goals \citep{colas2022autotelic}. For a more detailed description of these diversity levels, please refer to \ref{sec:appendix1}.

\subsection{The selected intrinsic reward algorithms}
\label{intrinsic rewards}
We augment the task reward with an intrinsic reward such that the total reward becomes: $r_{total}=r_{ext}+\beta * r_{int}$, where $r_{ext}$ is the extrinsic reward, $r_{int}$ is the intrinsic reward and $\beta$ is the intrinsic reward coefficient \citep{ladosz2022exploration}. % The best values of $\beta$ for each intrinsic reward method, refined via  hyperparameter search, are summarised in Table \ref{intrinsic reward coefficient} of Appendix \ref{sec:appendix4}.
The best-performing $\beta$ values, either sourced from the literature \citep{andres2022evaluation} or determined through a grid search (details provided in \ref{sec:appendix4}), are presented in Table \ref{intrinsic reward coefficient}, also located in \ref{sec:appendix4}.
%We tuned $\beta$ for each intrinsic reward method, with the best selected values summarised in Table \ref{intrinsic reward coefficient} of Appendix \ref{sec:appendix4}.
We select four different intrinsic reward methods, each representative of one of the four diversity levels:
\vspace{0.5em}

\noindent \textbf{State Count (State level diversity)} builds an intrinsic reward inversely proportional to the state visitation count \citep{strehl2008analysis}. For a transition \((s_t,a_t,s_{t+1})\), where $s_t$ is the current state, $a_t$ is the current action and $s_{t+1}$ is the next state, \(r_{int}(t)=1/\sqrt{N(s_{t+1})}\) with \(N(s_{t+1})\) being the number of times the state \(s_{t+1}\) has been visited so far during training. This algorithm considers only discrete, low dimensional state space. However, for RGB observations, where the state space is much larger and State Count is not feasible, we use SimHash \citep{tang2017exploration} to hash states before counting them. SimHash maps the pixel observations to hash codes according to the following equation, \review{with $h$ as the hashing function: $h(s_{t+1})= sgn(A* \phi(s_{t+1})) \in \{-1,1\}^k $. Here, $\phi$ is an embedding function,} $A$ is a matrix with i.i.d entries drawn from a standard normal distribution, $k$ is the sise of the hashed key, and $sgn(\cdot)$ maps a number to its sign. Then, the same intrinsic reward formula is applied but using the hashed observation: \review{ $r_{int}(t)=1/\sqrt{N(h(s_{t+1}))}$. }
\vspace{0.5em}

\noindent \textbf{Intrinsic Curiosity Module (ICM) (State + Dynamics level diversity)} uses curiosity as intrinsic reward. Curiosity is formulated as the error in the agent's ability to predict the outcome of its own actions in a learned state embedding space \citep{pathak2017curiosity}. Specifically, ICM trains a state embedding network, a forward and an inverse dynamic models. For a transition tuple \((s_t,a_t,s_{t+1})\), the embedding network $\phi: \mathcal{S} \rightarrow \mathcal{F}$ projects the current state \(s_t\) and next state \(s_{t+1}\) into the feature space $\mathcal{F}$ to get the embeddings \(\phi(s_t)\) and \(\phi(s_{t+1})\) respectively. Then, the inverse dynamics model $g: \mathcal{F} \times \mathcal{F}\rightarrow \mathcal{A}$ takes as input the current and next state embeddings, $\phi(s_t)$ and $\phi(s_{t+1})$ respectively, and predicts the action $a_t$ taken by the agent to move from state $s_t$ to state $s_{t+1}$. The state embedding network is updated, such that it only captures the features of the environment that are controlled by the agent's actions, and ignores the uncontrollable factors. The forward dynamics model $f: \mathcal{F} \times \mathcal{A} \rightarrow \mathcal{F}$ predicts the next state embedding $\phi(s_{t+1})$ given the current state embedding $\phi(s_t)$ and current action $a_t$. The intrinsic reward is the prediction error of the forward dynamics model: \(r_{int}(t)= \|{f(\phi(s_t), a_t)}-\phi(s_{t+1})\|^2_2\) \citep{pathak2017curiosity}.
\vspace{0.5em}

\noindent \textbf{Max Entropy RL (Policy level diversity)} augments the extrinsic reward with the policy entropy \(r_{int}(t)= H(\pi(.|s_t))\) to favour stochastic policies \citep{haarnoja2018soft,liu2019policy}. 
\vspace{0.5em}

\noindent \textbf{DIAYN (Skill level diversity)} aims to discover a set of diverse skills without supervision \citep{eysenbach2018diversity}. A skill is defined as a policy $\pi(a|s,z)$ conditioned on the state \(s\) and latent variable/ goal \(z\). DIAYN's objective is to maximise the mutual information between $z$ and every state in the trajectory generated by  $\pi(a|s,z)$. The intuition is to infer the skill from the state. At the start of each episode, a latent variable \(z\) is sampled from a uniform distribution \(p(z)\), then the agent acts according to that skill \(\pi(a|s,z)\) throughout the episode. \review{A discriminator $ q_\alpha (z|s)$} is trained to estimate the skill \(z\) from the state \(s\). \review{The intrinsic reward, defined by \(r_{int}(t)= log(q _\alpha (z|s_{t+1}))-log(p(z))\),} is used to push the agent for visiting states that are easily distinguishable in terms of skills. Then, the discriminator is updated to better predict the skill, and the policy is updated to maximise $r_{int}$ using any RL algorithm. % Note that DIAYN is an unsupervised skill discovery method, so unlike the intrinsic rewards above,  we do not combine DIAYN's intrinsic reward with the extrinsic reward. 
It is worth mentioning that DIAYN has been proposed as an unsupervised skill discovery method to favour robustness, fast adaptation to new tasks and hierarchical learning. Therefore, exploration is not the main goal of DIAYN. As a consequence, DIAYN's intrinsic reward can conflict with the agent's extrinsic reward, potentially jeopardising convergence if combined directly. To address this, we split the training budget between pretraining and finetuning phases. During pretraining, skills are learned using only intrinsic rewards. The learned weights are then used to initialise the policy and value networks for the finetuning phase with task-specific extrinsic rewards. The rationale for this approach is further explained in~\ref{DIAYN_extrinsic}, where we evaluate the performance of DIAYN when combined with extrinsic rewards.
\subsection{Environment}  \label{subsection:Environment}
We test on MiniGrid \citep{MinigridMiniworld23}, a widely used procedurally generated environment in RL exploration benchmarks \citep{raileanu2020ride,andres2022evaluation,wang2022revisiting} suitable to experiment with sparse rewards. We consider two types of observations: partially observable grid encodings, and partially observable RGB images (see~\ref{sec:appendix2}). The latter has a much larger state space, allowing us to investigate the scenarios challenging for State level diversity algorithms. To study the impact of different diversity levels on exploration, we select four environments with varying grid layouts and tasks, that highlight the strengths and weaknesses of various intrinsic reward methods: %Since we are interested in studying the impact of different diversity levels on exploration, we select the following four environments (having different grid layouts and tasks), which highlight the strengths/weaknesses of different intrinsic reward methods:

%\begin{enumerate}[leftmargin=0.5cm]
% \item \textit{Empty}: We choose this environment as control and it is the only one not procedurally generated (fixed initial and goal positions). It offers minimal constraints, allowing for freedom to solve the task in different ways. It is also interesting since it consists of a big, homogeneous room, which can exacerbate state aliasing: different MDP state, for example, different (x, y) position of the agents, conflate into the same observation \citep{mccallum1996reinforcement}. The mismatch is a source of error for state-counts.
% \item \textit{DoorKey}: This environment requires strategic exploration to locate keys and unlock doors. It stresses the importance of a trajectory to visit states in particular order. Methods that can learn skills and recognise these dependencies might perform better than state count-based methods, which treat all state visits equally without taking into account the order.  
% \item \textit{FourRooms}: This environment is characterised by the sparsity of rewards. The presence of multiple rooms encourages the agent to devise different strategies for navigation, fostering diversity.
% \item \textit{RedBlueDoors}: This environment also requires strategic exploration, but it is an easier task than DoorKey. It introduces colour-coded doors, requiring agents to exhibit high levels of cognitive flexibility. 
% \end{enumerate}
\begin{enumerate}
\item {\bf Empty}: We choose this environment as control and it is the only one not procedurally generated (fixed initial and goal positions). The setup imposes minimal constraints, providing freedom to solve the task in different ways. Consisting of one big homogeneous room, this environment is interesting since it can lead to state aliasing: different MDP states, for example, different $(x,y)$ positions of the agents, appear as identical observations \citep{mccallum1996reinforcement}.  This creates a challenge for state count-based methods, which count the observations they encounter and  therefore cannot differentiate between the true underlying states.
\vspace{1mm}
\newline 
\item {\bf DoorKey}: This environment requires strategic exploration to locate keys and unlock doors. It stresses the importance of a trajectory to visit states in particular order. Methods that can learn skills and recognise these dependencies might perform better than state count-based methods, which treat all state visits equally without taking into account the order.  
\vspace{1mm}
\newline 
\item {\bf FourRooms}: This environment is characterised by its sparsity of rewards. The presence of multiple rooms encourages the agent to devise different strategies for navigation, fostering the diversity in the trajectories or paths taken by the agent to achieve the goal.
\vspace{1mm}
\newline 
\item {\bf RedBlueDoors}: This environment also requires strategic exploration, but it is an easier task than DoorKey. It introduces colour-coded doors, requiring agents to exhibit high levels of cognitive flexibility. 
\end{enumerate}
 More details about the tasks, observation, and action spaces are included in~\ref{sec:appendix2}.

\subsection{Experimental Protocol}\label{subsection:Experimental_Protocol}  
We test the four algorithms in each environment, for each observation space. \review{We select Proximal Policy Optimisation (PPO) \citep{schulman2017proximal} as our baseline algorithm. PPO is a widely accepted and popular choice in RL research, known for its stability, robustness, and relatively high sample efficiency. Its simple implementation offers manageable computational costs, which enhances reproducibility and facilitates validation of results. Beyond its theoretical strengths, PPO has demonstrated success in complex real-world applications, such as large language model (LLM) research, underscoring its versatility and reliability for our study.}

We adopt the default hyperparameters from~\citep{andres2022evaluation}, listed in Table~\ref{PPO hyperparams} of~\ref{sec:appendix4}. This baseline algorithm comes with an entropy regularisation in the objective function to encourage a minimum level of exploration \citep{ahmed2019understanding}. Such regularisation is essential to avoid overfitting \citep{liu2019regularization}, and to stabilise the training process \citep{liu2019policy}. We set the entropy regularisation coefficient to $\epsilon=0.0005$ in all simulated algorithms. 
%We use the default hyperparameters of PPO from the previous work \cite{andres2022evaluation} (See Table \ref{PPO hyperparams} of Appendix \ref{hyperparams}). PPO (without any added intrinsic reward) serves as the baseline for our experiments. It usually has an entropy regularization in the objective function to encourage exploration \cite{ahmed2019understanding}, avoid overfitting \cite{liu2019regularization}, and stabilise the training process \cite{liu2019policy}. We fix the entropy regularization coefficient to the same value in the baseline and all the intrinsic reward methods: $\epsilon=0.0005$.
The selected value is large enough to guarantee a minimum level of stable convergence, but small enough to not affect our experiments. % Details of the network architectures are shown in Appendix \ref{sec:appendix3}.
%We also included, in Appendix \ref{hyperparams}, the best intrinsic reward coefficient $\beta$, which is a critical hyperparameter for all exploration methods.
We train each algorithm for $40 M$ frames on all environments. For DIAYN exclusively, we use $25M$ for pretraining and $15M$ for finetuning. Training curves, averaged over five runs with different seeds, are provided for all algorithms. \review{The simulations in this study were conducted on a high-performance computing node equipped with an NVIDIA TITAN X (Pascal) GPU featuring $12$ GB of VRAM, an Intel Xeon E5-2640 v4 CPU operating at $2.40$ GHz with $40$ cores, and $62$ GB of RAM. }
\begin{figure}[h]
\centering
\includegraphics[width=0.8\textwidth]{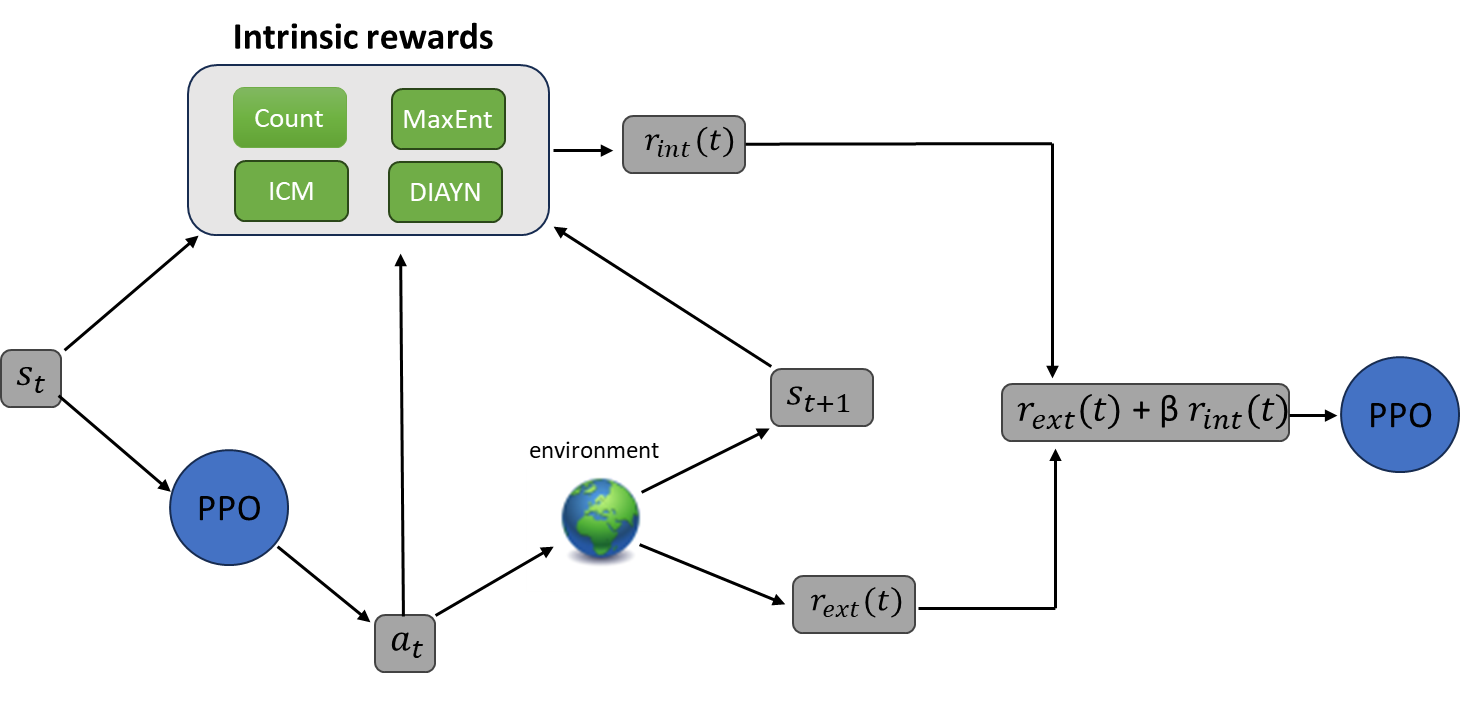}
\caption{Overview of the empirical study pipeline, illustrating the flow from input observations to action selection, and reward computation (both extrinsic and intrinsic) within the PPO framework.
}
\label{overall_pipeline}
\end{figure}

\review{\subsection{Model Architecture}\label{architecture}
 Figure~\ref{overall_pipeline} illustrates the pipeline of the empirical study, depicting the sequential flow of inputs, outputs, and reward computations within the model. At each time step \( t \), the input observation \( s_t \) (either a grid encoding or an RGB image) is processed by the PPO algorithm, which outputs an estimated policy and value function. The agent then takes an action \( a_t \) based on the estimated policy, transitions to the next state \( s_{t+1} \), and receives an extrinsic reward \( r_{ext}(t) \). Depending on the intrinsic reward method applied (State Count/SimHash, ICM, DIAYN, or Max Entropy), the agent computes an intrinsic reward \( r_{int}(t) \) for the transition \( (s_t, a_t, s_{t+1}) \), following the formulations in Section~\ref{intrinsic rewards}. The intrinsic and extrinsic rewards are combined $r_{total}(t)=r_{ext}(t)+\beta * r_{int}(t)$ and fed back into the actor-critic PPO network to refine the policy and value function. For DIAYN exclusively, we avoid combining intrinsic and extrinsic rewards, as discussed in Section~\ref{intrinsic rewards}.}

\begin{figure}[H]
  \centering
  \begin{subfigure}{0.5\linewidth}
    \centering
    \includegraphics[width=\linewidth]{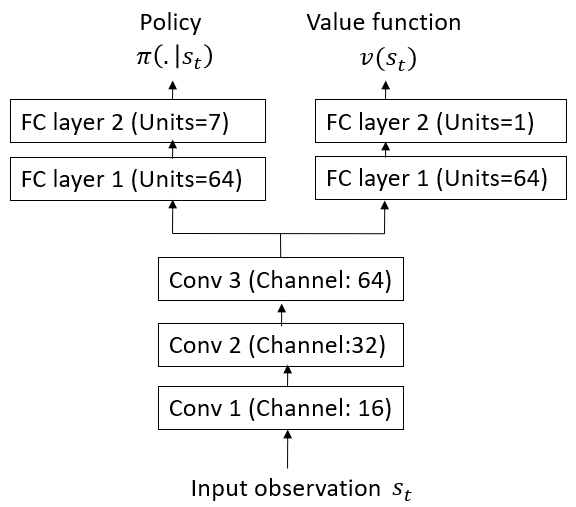}
    \caption{Actor-Critic network}
    \label{AC model}
  \end{subfigure}
  \hfill
  \begin{subfigure}{0.3\linewidth}
    \centering
    \includegraphics[width=\linewidth]{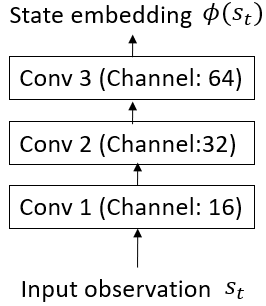}
    \caption{State Embedding}
    \label{SE model}
  \end{subfigure}
  \begin{subfigure}{0.32\linewidth}
    \centering
    \includegraphics[width=\linewidth]{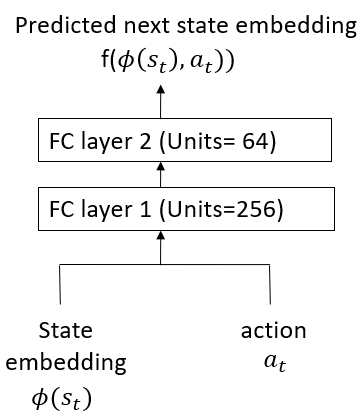}
    \caption{Forward dynamics network}
    \label{Fwd dynamics model}
  \end{subfigure}
  \hfill
  \begin{subfigure}{0.3\linewidth}
    \centering
    \includegraphics[width=\linewidth]{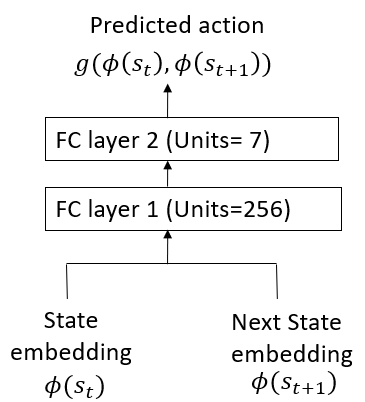}
    \caption{Inverse dynamics network}
    \label{Inverse dynamics model}
  \end{subfigure}
  \par\bigskip
  \begin{subfigure}{0.3\linewidth}
    \centering
    \includegraphics[width=\linewidth]{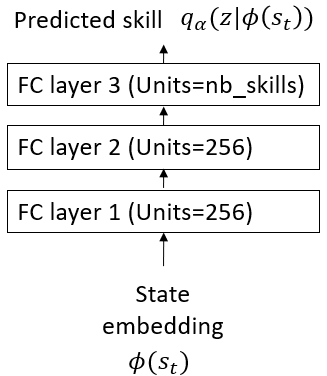}
    \caption{Discriminator network}
    \label{Dis}
  \end{subfigure}
  \caption{Neural Network Architectures}
  \label{NNs}
\end{figure}

\review{The Actor-Critic model architecture used in PPO employs a shared convolutional neural network (CNN) to process observations, which can be either grid encodings or RGB images. This CNN consists of three convolutional layers: the first layer has $16$ filters of size \(2 \times 2\) with ReLU activation, followed by a \(2 \times 2\) max-pooling layer; the second layer has $32$ filters of size \(2 \times 2\) with ReLU activation; and the third layer has $64$ filters of the same size and activation function. The CNN output then branches into two fully connected networks, designated as the actor and critic networks. Each network includes a hidden layer with 64 units and Tanh activation. The actor network produces action probabilities, while the critic network outputs the value function. Figure \ref{AC model} provides an overview of this architecture.}

\review{The PPO architecture remains consistent across all intrinsic reward methods. Some methods, however, require additional auxiliary networks, such as the embedding networks \(\phi\) for ICM, DIAYN, and SimHash (see Figure \ref{SE model}), forward (\(f\)) and inverse (\(g\)) dynamics networks in ICM (Figures~\ref{Fwd dynamics model} and~\ref{Inverse dynamics model}), and the discriminator network \(q_\alpha\) in DIAYN (Figure~\ref{Dis}). For ICM and DIAYN, the state embedding network follows the same CNN architecture as PPO to extract features from observations (see Figure \ref{SE model}). SimHash further appends a fully connected layer to the embedding network, reducing the RGB image embedding to a $512$-dimensional vector prior to hashing. }

\vspace{1.5em} % Adjust the value as needed
\subsection{Evaluation Metrics} \label{metrics}
We analyse each of the intrinsic rewards, according to five metrics:

\noindent{\bf Episodic return:} We plot the episodic extrinsic return averaged over the number of parallel actors $ \frac{1}{N} \sum_{n=1}^{n=N} \sum_{t=1}^{t=H} r_t^n$, where $N$ is the number of actors and $H$ is the length of the episode. This metric captures the convergence speed and learning ability of the intrinsic reward method. 
\vspace{1mm}
\newline 
\noindent{\bf Observation coverage:} This metric offers insight into the extent of exploring the observation space. We count how many unique observations (grid encodings or RGB) have been visited during training. We normalise this metric over the highest coverage achieved by the intrinsic reward methods. Observation coverage mirroring state coverage suggests that the neural network's embedding captures historical information and effectively represents the state.
\vspace{1mm}
\newline 
\noindent{\bf Agent's position coverage:} This metric shows the proportion of $(x,y)$ grid positions visited by the agent so far during training: $N_{visited}(x,y)/N_{total}(x,y)$. $N_{total}(x,y)$ is the number of all possible grid positions that the agent can visit. This metric captures how well the agent has explored the position space, which is different from the observation space in a partially observable framework.
\vspace{1mm}
\newline 
\noindent{\bf Policy Entropy:} This metric evaluates the level of stochasticity of the policy. %Higher entropy suggests that the policy is diverse.
We use the average Shannon entropy: $H = - \frac{1}{T} \sum_{i=1}^{T} \sum_{j=1}^{7} p(a_j) \log p(a_j)$ where $T$ is the number of frames.
\vspace{1mm}
\newline 
\noindent{\bf Time steps of the first, second and third reward discoveries:} We record the number of frames at which the sparse reward is found by each intrinsic reward for the first, second and third times. Note that ``number of frames'' refer to the number of times the agent interacted with the environment throughout the training. This metric sheds light on the speed and effectiveness of the exploration method to discover the high reward states, as well as learning to revisit these states.

 Finally, we include further visualisations (heatmaps) of the state visitation count ($(x,y)$ positions) in \ref{sec:appendix5}. These heatmaps represent the proportion of visits to each grid position $(x,y)$ relative to the total number of frames. To generate them, we train the agent for $10M$ frames in singleton environments, where the maze layout remains fixed across training episodes. This setup highlights the agent's exploration patterns on a consistent grid map. \review{Figures~\ref{heatmap_Empty_NonRGB},~\ref{heatmap_DoorKey_NonRGB}, ~\ref{FouRooms_NonRGB}, and ~\ref{RedBlueDoors_NonRGB} in~\ref{sec:appendix_NonRGB_results} display results for grid-encoded observations, while Figures~\ref{heatmaps_Empty_RGB},~\ref{heatmaps_DoorKey_RBG},~\ref{FouRooms_RGB}, and ~\ref{RedBlueDoors_RGB} in~\ref{sec:appendix_RGB_results} show results for RGB observations. These visualisations illustrate the areas of the grid explored by the agent during training across the four environments: Empty, DoorKey, FourRooms, and RedBlueDoors.}
 % Images created from https://colab.research.google.com/drive/1QiyDbjhjL5F3dbs-uFLrct6Yxu3w2Mrs?usp=sharing
\begin{figure}[h]
\centering
\includegraphics[width=\textwidth]{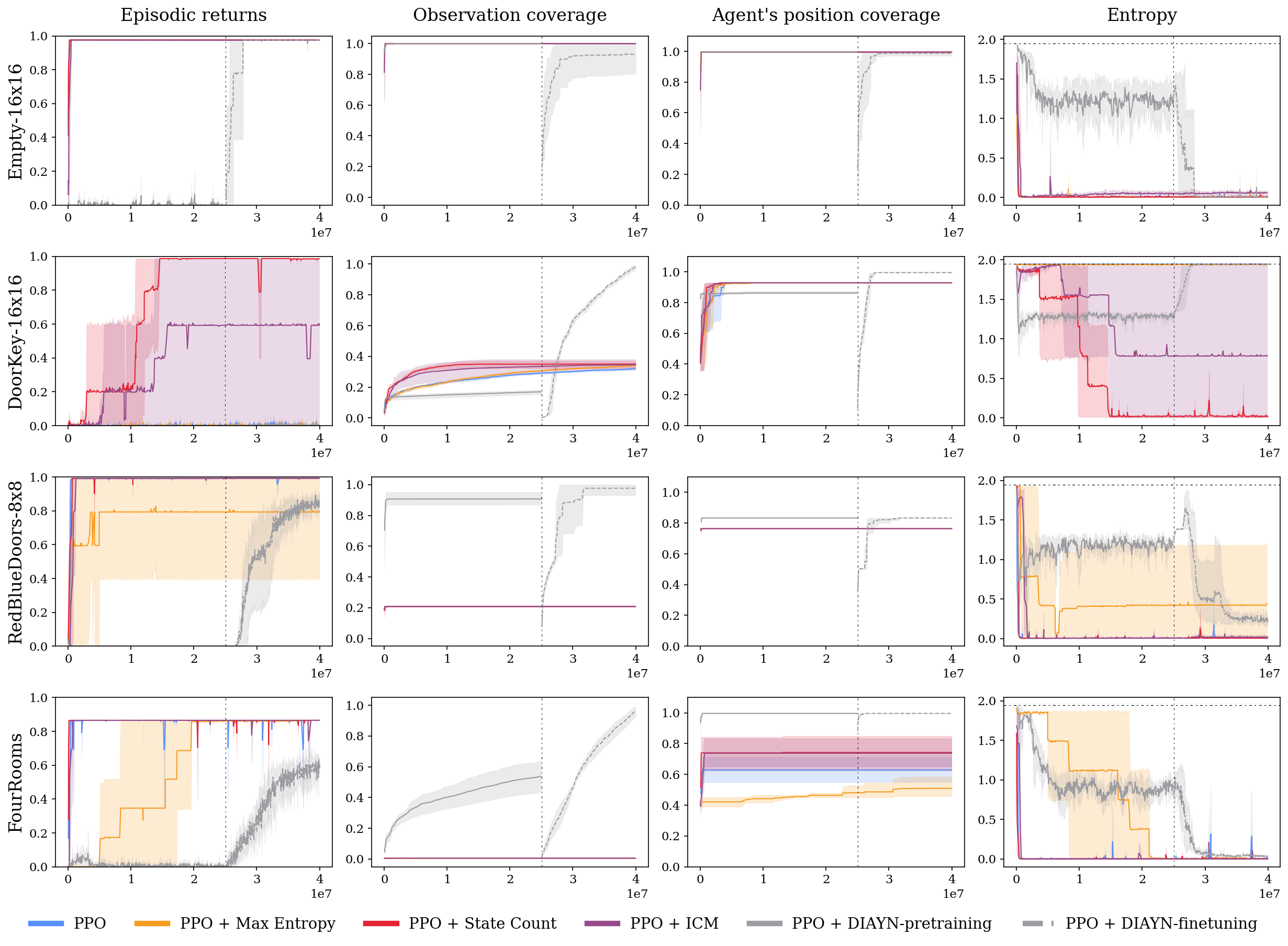}
\caption{
\review{Performance of four metrics -- Episodic Return, Observation Coverage, Agent's Position Coverage, and Entropy -- plotted against the number of transitions (frames) processed by the environment. Observations are represented as grid encodings. The results, averaged over five seeds with standard deviation shading, include evaluations across the four environments described in Section~\ref{subsection:Environment}. The baseline model, PPO, operates without intrinsic rewards, while the other four algorithms incorporate intrinsic rewards detailed in Section~\ref{intrinsic rewards}. For DIAYN, we differentiate between pretraining (for frames \( < 25M \)) and finetuning (for frames \( \in [25M, 40M] \)). Vertical dash-dot lines indicate the beginning of DIAYN finetuning, while horizontal dash-dot lines represent the theoretical maximum entropy of the policy, defined as \( H_{max} = H(\mathcal{U}_{|A|}) = \log(|A|) \).}
}
\label{MatrixA}
\end{figure}

\begin{figure}[h]
\centering
\includegraphics[width=\textwidth]{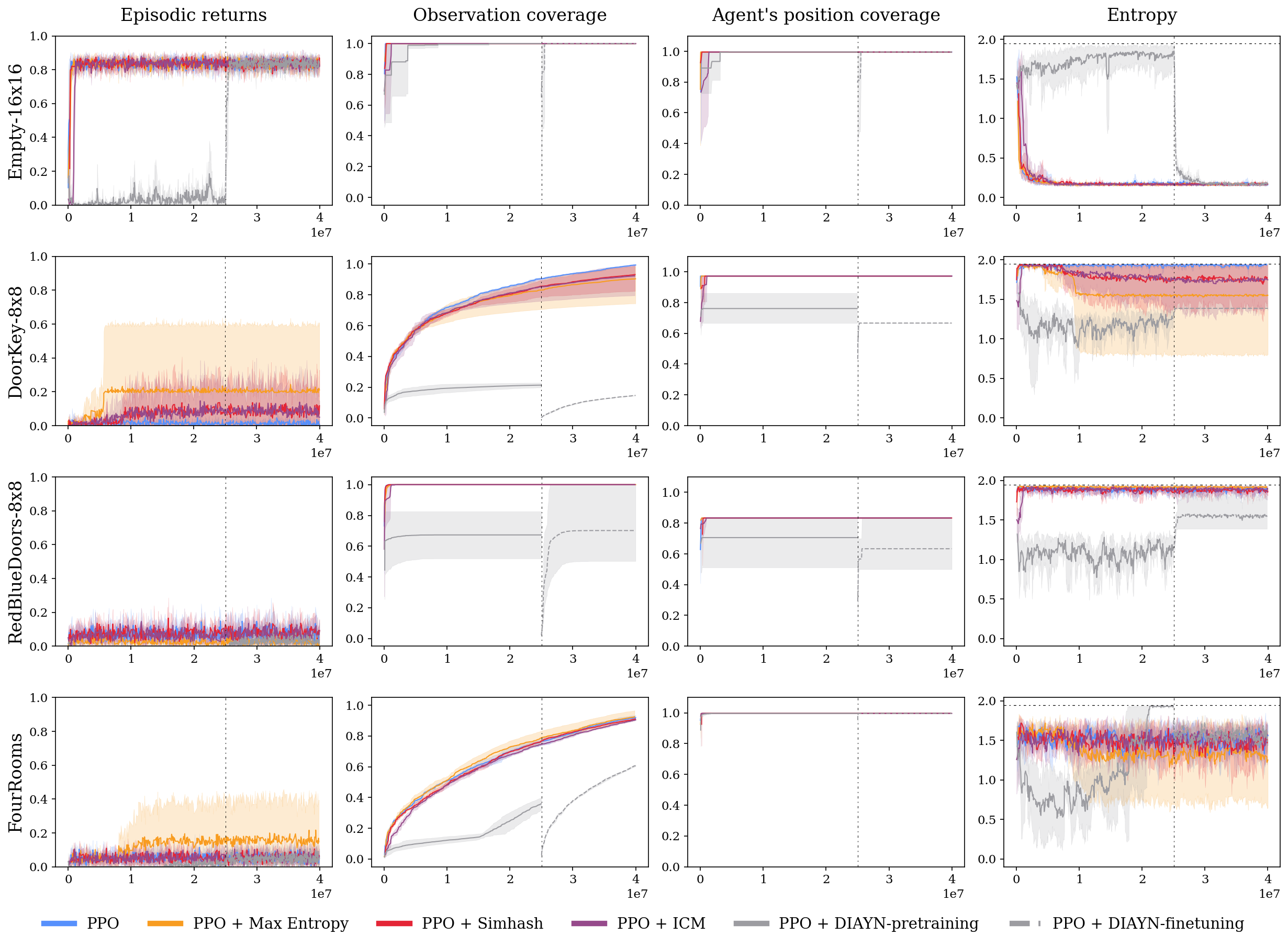}
\caption{
Analogous to Figure \ref{MatrixA}, but observations are partial RGB images.
}
\label{MatrixB}
\end{figure}

\section{Experimental results and discussion}

We discuss the following three questions to analyse the performance of the exploration algorithms:
\begin{itemize} [leftmargin=0.5cm]
    \item [--] RQ1: Do different intrinsic rewards lead to different return performance/sample efficiency for both grid encodings and RGB partial observations?
    \item [--] RQ2: What are the characteristics (strengths/weaknesses) of each intrinsic reward method, and what are the practical recommendations to select intrinsic rewards?
    \item [--] RQ3: How different intrinsic rewards impact efficiency in discovering the sparse reward? Is there any link with credit assignment?
\end{itemize}

\subsection{RQ1: Return performance of the different intrinsic rewards}

In terms of episodic return, State Count has the best performance with low-dimensional observations (grid encodings) on all environments (See column $1$ of Figure \ref{MatrixA}). It converges to the maximum return with the least number of frames. In the case of DoorKey 16x16, where many algorithms, including PPO, Max Entropy, and DIAYN, struggle to solve the task, State Count emerges as the top performer, successfully obtaining the key and attaining the highest return. Following closely, ICM demonstrates lower sample efficiency. However, this is not the case for RGB observations (Refer to column $1$ of Figure \ref{MatrixB}), in which SimHash (equivalent to State Count) performs poorly on most environments. The failure of SimHash in the case of RGB observations can be attributed to the challenge in adequately representing the significant features present in the high-dimensional states. \review{RGB images contain an abundance of extraneous pixel-level details that are irrelevant to the task, requiring the agent to represent only the meaningful features.
SimHash, which uses a simple hashing mechanism to represent states, struggles to capture the essential features in RGB states due to their sparse and coarse encoding mechanisms. This limitation becomes especially apparent in environments requiring a high level of feature abstraction and attention to object relationships, such as DoorKey 16x16, where misrepresenting critical details hinders the agent’s ability to navigate effectively.}

Max Entropy is less impacted by such representation learning difficulties. It manages to have a slightly higher return on DoorKey 8x8 and FourRooms environments in the case of RGB observations (Figure \ref{MatrixB}). \review{This robustness can be attributed to Max Entropy's tendency to encourage diverse policy exploration without heavily relying on specific state representations, which provides a certain level of resilience to noisy feature extraction.} All other intrinsic rewards struggle to solve the tasks (except for Empty 16x16) and consistently maintain a high level of non-decreasing policy entropy. %This is understandable, as the  agent must extract features directly from partially observable raw images, which is more challenging.
\review{This is likely because these methods rely on high-quality state representations to produce meaningful novelty signals. In high-dimensional RGB observations, however, they tend to generate less informative intrinsic rewards. This results in difficulty differentiating between truly novel states and irrelevant pixel-level variations, causing policy learning to stagnate.}

DIAYN finetuning has a worse average return compared to the baseline PPO, in both grid encodings and RGB scenarios. This shows that initialising the actor-critic weights with DIAYN skills does not improve sample efficiency compared to random initialisation. Note that DIAYN pretraining does not collect any extrinsic reward, because it is trained to maximise the intrinsic reward generated by the discriminator and not the true task reward. \review{ We hypothesise that the limited skill label space, compared to the vast state space, promotes the learning of static skills that lack adaptability and fail to transfer effectively to the target task.
Specifically, the states encountered by different skills tend to vary only slightly, enabling skill differentiation but not necessarily the development of semantically meaningful or broadly transferable skills. }

\subsection{RQ2: Characteristics of each intrinsic reward algorithm}

\textbf{State Count / SimHash} demonstrates the best sample efficiency in grid encodings, enabling efficient task-solving in small state/action spaces. Additionally, it ensures a fast coverage of observations and grid positions compared to other algorithms, depicted \review{in columns $2$ and $3$ of Figures \ref{MatrixA} and \ref{MatrixB}.} Furthermore, as it converges to the optimal policy, it exhibits a fast decreasing policy entropy due to the diminishing intrinsic reward effect with increasing states count. Examination of the heatmaps (\ref{sec:appendix5}) reveals that State Count offers the most uniform coverage of the state space across all environments. This  enables the algorithm to identify the optimal path, while maintaining a balanced approach between exploration and reward maximisation. Remarkably, in the DoorKey environment (Figure \ref{heatmap_DoorKey_NonRGB} in~\ref{sec:appendix5}), State Count demonstrates a tendency to revisit the area around the key more frequently. However, despite these strengths, a notable limit arises in its inability to effectively handle RGB images. In such cases, the algorithm struggles to accurately count or represent pixels, thereby limiting its applicability in scenarios of high-dimensional state spaces. %In the case of good state representations, State Count could be applied  in the observation domain, but this might still be affected by the challenge of learning good representations \citep{henaff2022exploration}.
%One solution could be to learn good state representations, then counting those instead of the observations themselves, but it is not easy to design a feature extractor that excels across all tasks \citep{henaff2022exploration}.
As a practical recommendation, State Count is a good choice for small, discrete environments, but struggles with complex, high-dimensional ones. \review{%Count-based methods are not scalable in continuous and large state spaces.
Although we did not explore how different representations impact the performance of State Count in this study, incorporating representation learning techniques presents an interesting avenue for future research.}
\vspace{0.5em}

\noindent \textbf{Intrinsic Curiosity Module (ICM)} exhibits favourable return performance and effectively explores the observation and position spaces, similarly to State Count, as they both prioritise exploration within the state space. In environments characterised by low-dimensional state spaces (Figure \ref{MatrixA}), ICM showcases consistent stability in solving tasks across diverse scenarios. However, ICM’s convergence speed generally lags behind State Count due to the added computational complexity of training both forward and inverse dynamics models. This additional overhead likely introduces inefficiencies that slow down exploration, as shown in heatmap analyses (~\ref{sec:appendix5}), where ICM's slower rate of grid position exploration is evident. These heatmaps illustrate that while ICM achieves thorough state coverage, it does so at a slower rate, potentially limiting its efficiency in tasks requiring rapid convergence.
%However, its convergence speed typically lags behind that of State Count. This can be attributed to the higher computational complexity associated with training both forward and inverse dynamics models. This discrepancy is further evidenced by heatmap analyses (Appendix \ref{sec:appendix5}), revealing ICM's comparatively slower exploration of grid positions.
Moreover, similarly to State Count, ICM encounters challenges in effectively processing RGB images (Figure \ref{MatrixB}). \review{The pixel-based inputs add significant complexity, making it difficult for ICM’s dynamics models to effectively process and encode meaningful features. This limitation suggests that ICM’s performance may be hampered in visually complex environments.}
\vspace{0.5em}

\noindent \textbf{Max Entropy} can solve most environments in the case of grid encoding observations (except DoorKey) (Figure \ref{MatrixA}). However, it does not converge faster than State Count and shows a slightly lower average return because it fails for some of the runs (such as on RedBlueDoors). \review{ This instability arises from the algorithm's tendency to promote high stochasticity in the policy, even when a more deterministic approach would suffice, ultimately affecting the average performance. }%This reflects the instability of the algorithm, which pushes for high stochasticity in the policy, even when it is not needed, therefore affecting the average performance.
By analysing the heatmaps, we can see that Max Entropy explored unnecessarily \review{ or became confined to certain regions of the state space,} especially in easy environments such as FourRooms and RedBlueDoors (Figures \ref{FouRooms_NonRGB} and \ref{RedBlueDoors_NonRGB} in~\ref{sec:appendix5}). %This explains why Max Entropy takes longer to converge to the optimal path compared to ICM. %We deduce that Max Entropy  pushes the agent to try out all possible actions including unused ones, which can distract the agent from finding the goal sooner. intends to direct the policy towards the direction of high future entropy
\review{This unnecessary exploration delays convergence to optimal paths, as the agent is distracted from effectively reaching the goal. The algorithm’s inclination to prompt the agent to try all possible actions, including those rarely relevant to task success, can divert focus and hinder progress. Additionally, a drawback of the Max Entropy approach is that states with lower entropy may be visited less frequently or even overlooked. As discussed by~\citet{han2021max}, the maximum entropy strategy, which optimises policies to reach high-entropy states, does not always foster effective exploration. Rather, it can create positive feedback loops where the agent becomes overly focused on high-entropy areas, limiting its ability to comprehensively explore the environment. This might reduce the likelihood of reaching less-visited yet potentially critical states.}

%Another drawback of Max Entropy, is that it may cause that the states with low entropy are rarely accessed. As denoted by~\citet{han2021max}, maximizing entropy by optimizing policies to reach high-entropy states in the future does not always lead to effective exploration. Instead, it can create positive feedback loops that actually impede exploration.}
%\noindent As a pratical recommendation, in grid-like environments with high dimensional action space (where many actions could be unused), Max Entropy is not ideal as an exploration method, however, in the case of large state space, Max Entropy is not bad. 
%As a practical recommendation, in grid-like environments with a large number of unused actions, Max Entropy may not be a well-suited exploration method.  % Thus, as a practical recommendation in grid-like environments, for large action spaces with many unused actions, Max Entropy exploration is not ideal.
Nevertheless, in the case of partial RGB observations (Figure \ref{MatrixB}), Max Entropy is less impacted by representation learning challenges. We observe that on  DoorKey and FourRooms, it slightly outperforms SimHash in terms of return and shows a decrease in the policy entropy (see columns $1$ and $4$ of Figure \ref{MatrixB}), as it succeeds in reaching the goal in several runs. Therefore, for grid-based settings with high-dimensional state spaces, where simply counting states becomes impractical, maximising entropy can be a valuable alternative exploration strategy to State Count.  
%pushing for Policy level diversity is recommended in high-dimensional state spaces, where State Count fails.
As a practical recommendation, Max Entropy may not be the most effective exploration method in grid-like environments with high-dimensional action spaces, where many actions are unused. However, it performs adequately in environments with large state spaces and small action spaces.
\vspace{0.5em}

\noindent \textbf{DIAYN} has generally the worst average return compared to the other three intrinsic rewards in both grid encodings and RGB scenarios. This is attributed to the tradeoff between the ability to discriminate between different skills and optimality. \review{The need to generate distinguishable skills often leads DIAYN to prioritise visits to easily discriminable states over achieving optimal exploration.} In the case of low-dimensional state space (Figure \ref{MatrixA}), it is surprising that DIAYN finetuning has the highest observation and position coverages on most environments (DoorKey, FourRooms and RedBlueDoors).
The ease of discriminating observations (due to distinct grid encodings reflecting different object types, colours, or status) drives DIAYN to prioritise visiting them. Unlike environments with distinct features, DIAYN struggles to cover the observation space in Empty 16x16 due to the difficulty of discriminating observations in a near-uniform grid (mostly walls). \review{This further underscores DIAYN’s reliance on environments with clear, discriminable features for effective exploration.} Moreover, the poor state space coverage by DIAYN (both pre-trained and fine-tuned) in the RGB setting (Figure \ref{MatrixB}) indicates limitations in the discriminator's ability to discriminate between RGB observations. This suggests that the additional challenge of representation learning exacerbates the discriminator's learning difficulties. \review{The presence of high-dimensional visual data introduces an added layer of complexity as the agent must learn both to distinguish visual features and navigate the space effectively.}
By further analysing the exploration pattern of DIAYN through the heatmaps, we notice the following:
DIAYN demonstrates uneven state coverage, often focusing on corner areas or becoming restricted to specific regions within the grid that contain easily distinguishable states (For example, see Figures \ref{heatmap_DoorKey_NonRGB} \ref{RedBlueDoors_NonRGB} \ref{FouRooms_RGB} \ref{RedBlueDoors_RGB} in~\ref{sec:appendix5}). This suggests potential limitations in its ability to explore diverse regions and acquire transferable skills. Without reaching different target positions (e.g., door/key/goal), the skills lack meaningful variations and adaptability. \review{We hypothesise that this is due to DIAYN's mutual information (MI) objective, which does not explicitly maximise the entropy of the state distribution ~\citep{yang2023behavior} and does not promote broad state coverage~\citep{tolguenec2024exploration}. The agent tends to receive higher rewards for visiting known states rather than exploring novel ones, as fully discriminable states yield a high MI reward \citep{kim2024learning}. This can hinder novel state exploration, and discourage the agent from
learning far-reaching skills ~\citep{yang2023behavior}. Consequently, DIAYN might potentially constrain the diversity of learned skills to those that are easier to distinguish, but not necessarily effective for broad exploration or task relevance.} In our particular setting, DIAYN also encounters difficulty in learning the abstract skill space effectively. This challenge might be particularly pronounced due to partial observability. As a practical recommendation, learning unsupervised skills with DIAYN does not help exploration in MiniGrid framework, especially in strategical tasks. Nevertheless, pushing for diversity of skills can be useful for skill-chaining, fast adaptation to environment changes, robustness, and generalisation to different tasks. We emphasise that our results hold only for our particular setting where skill learning turns out to be antagonist to exploration and sample efficiency in MiniGrid. This might not hold in other environments that could benefit from such skills to converge faster. 
\review{It is also worth noting that exploring factors such as the skill space, the initial skill distribution, and incorporating state abstraction techniques, along with auxiliary exploration mechanisms to enhance state coverage of skills, could significantly improve DIAYN’s performance. However, because this constitutes a substantial variation from the original algorithm, we leave these considerations for future work.}

\begin{figure}[h]
    \centering
    \includegraphics[width=\textwidth]{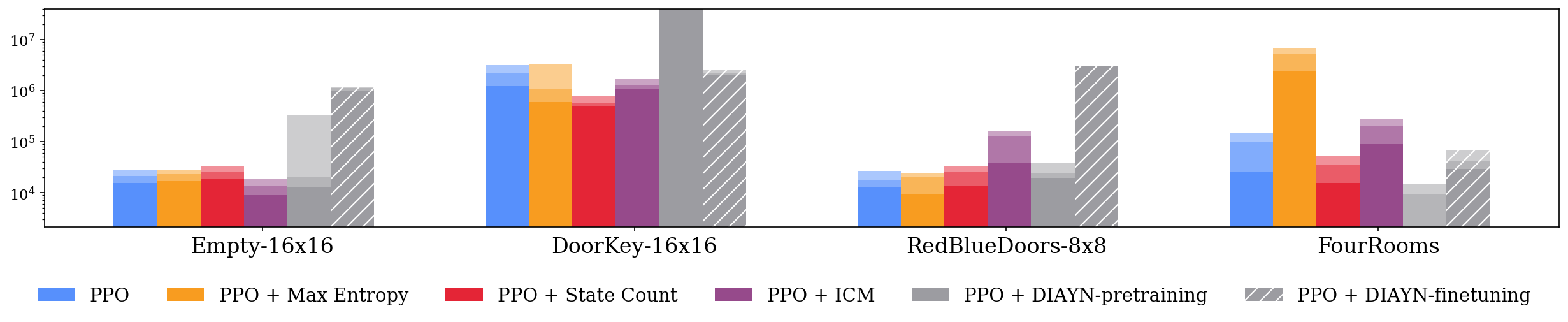}
    \caption{\review{Histogram showing the average number of frames required for each exploration method (PPO baseline and the four intrinsic rewards detailed in Section~\ref{intrinsic rewards}) to discover rewards across the environments described in Section~\ref{subsection:Environment}. Observations are grid encodings. Each bar is divided into three progressively fading compartments, representing the frames at which the first, second, and third rewards are collected during training, with lower values indicating better performance. Results are averaged over five runs. For variation measures alongside average results, see the tables in~\ref{sec:appendix5}.}}
    \label{fig:results:sym-rewards}
\end{figure}

\begin{figure}[h]
    \centering
    \includegraphics[width=\textwidth]{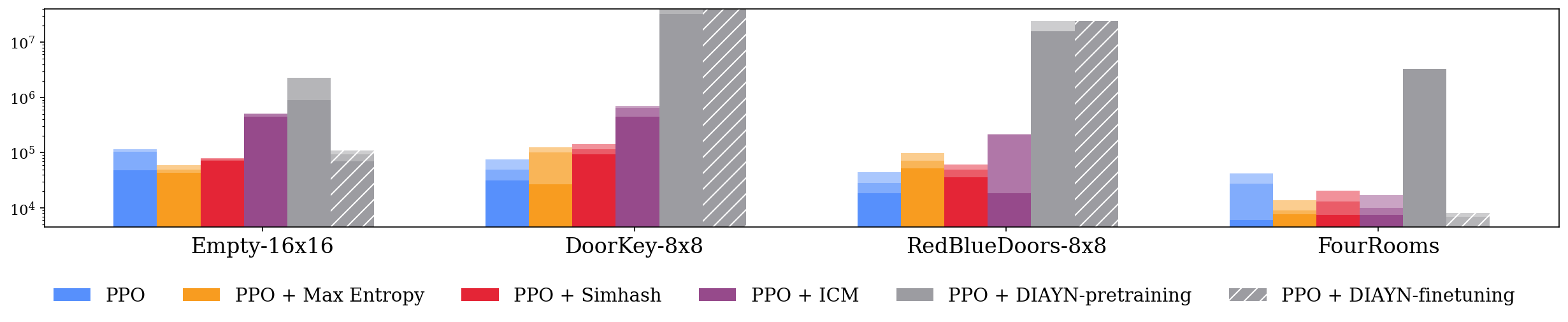}
    \caption{Analogous to Figure \ref{fig:results:sym-rewards} but observations are partial RGB images. }
    \label{fig:results:rgb-rewards}
\end{figure}

\subsection{RQ3: First, second and third instances of discovering the sparse reward }
We record the timesteps at which the sparse reward is found by each of the intrinsic rewards for the first, second and third times in both grid encodings (Figure \ref{fig:results:sym-rewards}) and RGB (Figure \ref{fig:results:rgb-rewards}) scenarios. \review{For more detailed results, including averages and standard deviations, refer to Tables~\ref{table_empty_nonRGB}, ~\ref{table_doorkey}, ~\ref{table_redbluedoors_nonRGB}, and ~\ref{table_fourRooms_nonRGB} in~\ref{sec:appendix_NonRGB_results}, as well as Tables~\ref{table_empty_RGB}, ~\ref{table_Doorkey_RGB}, ~\ref{table_RedBlueDoors_RGB}, and ~\ref{table_FourRooms_RGB} in~\ref{sec:appendix_RGB_results}.} 
We notice that in the case of low-dimensional observation space, State Count  (which has the highest return performance) finds the reward soon on most environments, while DIAYN takes time to reach the goal, especially on strategical tasks. For example, on DoorKey environment, which represents a strategical task, State Count is the first intrinsic reward to find the task reward, while DIAYN finetuning is the last and DIAYN pretraining does not reach the goal at all within the pretraining time. This shows that DIAYN exhibits limitations in acquiring skills that achieve the goal sequence of visiting the key, the yellow door, and the green goal in this specified order. \review{This limitation is likely due to DIAYN’s focus on skill diversity rather than directed exploration, making it less effective in tasks requiring structured sequences.} Another interesting observation is that the algorithm that discovers the reward the fastest for the first time, might not be the fastest in visiting the rewarding state a second and third time. This implies that diversity impacts credit assignment.
%Another interesting insight is that different exploration algorithms succeed to reach the goal at different times and rates, which affects credit assignment. 
For example, on DoorKey (see Table \ref{table_doorkey} in~\ref{sec:appendix5}), Max Entropy finds the first reward before ICM for the first time, but it takes more time to learn that it should go back to the reward for the third time. \review{This is paramount to design a sample efficient algorithm because visiting rewards more often provides more informative learning signals and allows to learn credit faster, more accurately and with less variance \cite{pignatelli2024a}. }\review{This implies that although Max Entropy promotes policy exploration, it may lack mechanisms for prioritising or remembering rewarding states that consistently provide useful learning signals.}

In contrast, the results vary across other environments, highlighting how different algorithms perform under varying conditions. Notably, in the FourRooms environment (Figure \ref{fig:results:sym-rewards}),  DIAYN pretraining is the first to find the reward, as opposed to the case of strategical tasks. In an environment consisting of several identical compartments, learning skills could lead to quick reward discovery even though it does not directly maximise the task reward. \review{This suggests that DIAYN might be advantageous in environments with structural similarity, where learned skills can be reused across similar compartments. }%This is likely because the learned skills are transferable across compartments. 
For the case of RGB observations (Figure \ref{fig:results:rgb-rewards}), we observe that PPO and Max Entropy are among the fastest methods to find the reward on most environments, surpassing SimHash. This reinforces that hypothesis that, when scaling to high observation spaces, entropy might be a better strategy to push for exploration rather than counting states. %rather than pushing for diversity on the State level.
DIAYN finetuning also takes long time to find the reward, especially for strategical tasks such as DoorKey and RedBlueDoors\review{, reinforcing that DIAYN’s emphasis on diversity may limit its ability to prioritise reaching task-relevant states in complex, sequential tasks. Integrating the mutual information (MI) objective of DIAYN with trajectory-based metrics between states to enhance exploration could be a potential direction for handling strategic tasks.}

\section{Conclusions}
In this work, we have re-interpreted intrinsic reward techniques in the literature using a diversity perspective (State, State + Dynamics, Policy and Skill levels of diversity). We conducted empirical studies on MiniGrid, to understand the differences between these diversity levels in a partially observable and procedurally generated framework.

We found that the homogeneity of the state coverage imposed by State Count (representing State level diversity) %diversity on the State level (represented by State Count)
has led to the best sample efficiency on many MiniGrid tasks. State level diversity improves the convergence speed in strategical tasks, covers well the state space and leads to a fast decrease of policy entropy and intrinsic reward. This decreasing rate of the intrinsic reward aligns well with finding the optimal policy which avoided the dominance of the intrinsic reward. However, State level diversity is fragile and requires good state representations, while entropy maximisation seems to be slightly more robust when dealing with image-based observations. Learning good state representations is challenging, so entropy maximisation is a practical alternative. 
Moreover, DIAYN (representing Skill level diversity) struggles with exploration in MiniGrid due to the difficulty of learning the skill space and exploring within it, in a procedurally generated partially observable setting. 
\review{
\subsection{Limitations and Future works}
This study serves as an initial exploration into the relationship between exploration and diversity imposed by intrinsic rewards. While we provide insights into this relationship, several limitations remain to be addressed in future work.}

\review{Firstly, we examine only one representative intrinsic reward method for each level of diversity. This choice may not capture the full range of behaviours within each category, potentially limiting the generalisability of our findings. Expanding this work to benchmark a broader selection of intrinsic reward methods would improve the applicability of our results.}

\review{Additionally, the effectiveness of intrinsic rewards is closely tied to the environment in which they are applied. Our experiments are restricted to the MiniGrid environment, specifically using grid encodings and RGB observations. Future studies could benefit from exploring more complex and varied environments, such as Mujoco~\citep{todorov2012mujoco}, Atari~\citep{bellemare2013arcade}, MiniHack~\citep{samvelyan2021minihack}, and MiniMax (Autocurricula)~\citep{jiang2023minimax}, where the impacts of different diversity levels might yield more distinct behaviours. Some intrinsic reward methods may excel in certain environments but perform poorly in others. Thus, identifying conditions under which each intrinsic reward method performs best across diverse environments would be a valuable contribution.}

\review{Moreover, while diversity can enhance exploration, it may also impede performance  as discussed in \citep{lin2024curse} in a phenomenon names \emph{the curse of diversity}. Therefore, pinpointing the conditions under which diversity aids rather than hinders performance—or developing strategies to counterbalance the potential negative effects of diversity—remains an open research question.}

\review{For the competence-based category, we employed DIAYN, a method that learns a skill space autonomously. Other goal-conditioned approaches, such as those learning different goal representations~\citep{pong2019skew}, predefining goal abstractions~\citep{colas2019curious}, or employing careful skill composition/chaining, may yield more efficient exploration strategies. Investigating these approaches could offer further insights into competence-based intrinsic rewards.}

\review{Finally, representation learning -- especially as applied in conjunction with intrinsic reward methods -- also significantly impacts exploration efficacy. Analysing how representation learning interacts with different levels of diversity and affects exploration performance is an important direction for future research.}

\review{Another interesting future work could explore integrating intrinsic reward diversity with context-aware RL~\citep{lin2022context, kiani2023novel,thakoor2013context}. This integration could improve exploration strategies by adapting them to specific environmental or task contexts. However, this is beyond the scope of the current work, which focuses solely on intrinsic rewards without additional adaptive mechanisms.}

\section*{Declarations}
%The authors declare no conflicts of interest, confirm that ethical approval is not required, and note that there is no data associated with this manuscript.
\textbf{Conflict of interests:} The authors have no relevant financial or nonfinancial interests to disclose.

\noindent \textbf{Ethical approval:} Not applicable.

\noindent \textbf{Data availability:} This manuscript has no associated data.

% \subsubsection*{Broader Impact Statement}
% \label{sec:broaderImpact}
% In this optional section, RLC encourages authors to discuss possible repercussions of their work, notably any potential negative impact that a user of this research should be aware of. 

% \label{sec:ack}
% Use unnumbered third level headings for the acknowledgments. All acknowledgments, including those to funding agencies, go at the end of the paper. Only add this information once your submission is accepted and deanonymised. 

%%%%%%%%%%%%%%%%%%%%%%%%%%%%%%%%%%%%%%%%%%%%%%%%%%%%%%%%%%%%%%%%
%% Appendices
%%%%%%%%%%%%%%%%%%%%%%%%%%%%%%%%%%%%%%%%%%%%%%%%%%%%%%%%%%%%%%%%

\appendix
\renewcommand{\thesection}{\appendixname~\Alph{section}}

\section{Diversity levels Categorisation}
\begin{figure}[h]
\centering
\includegraphics[width=\textwidth]{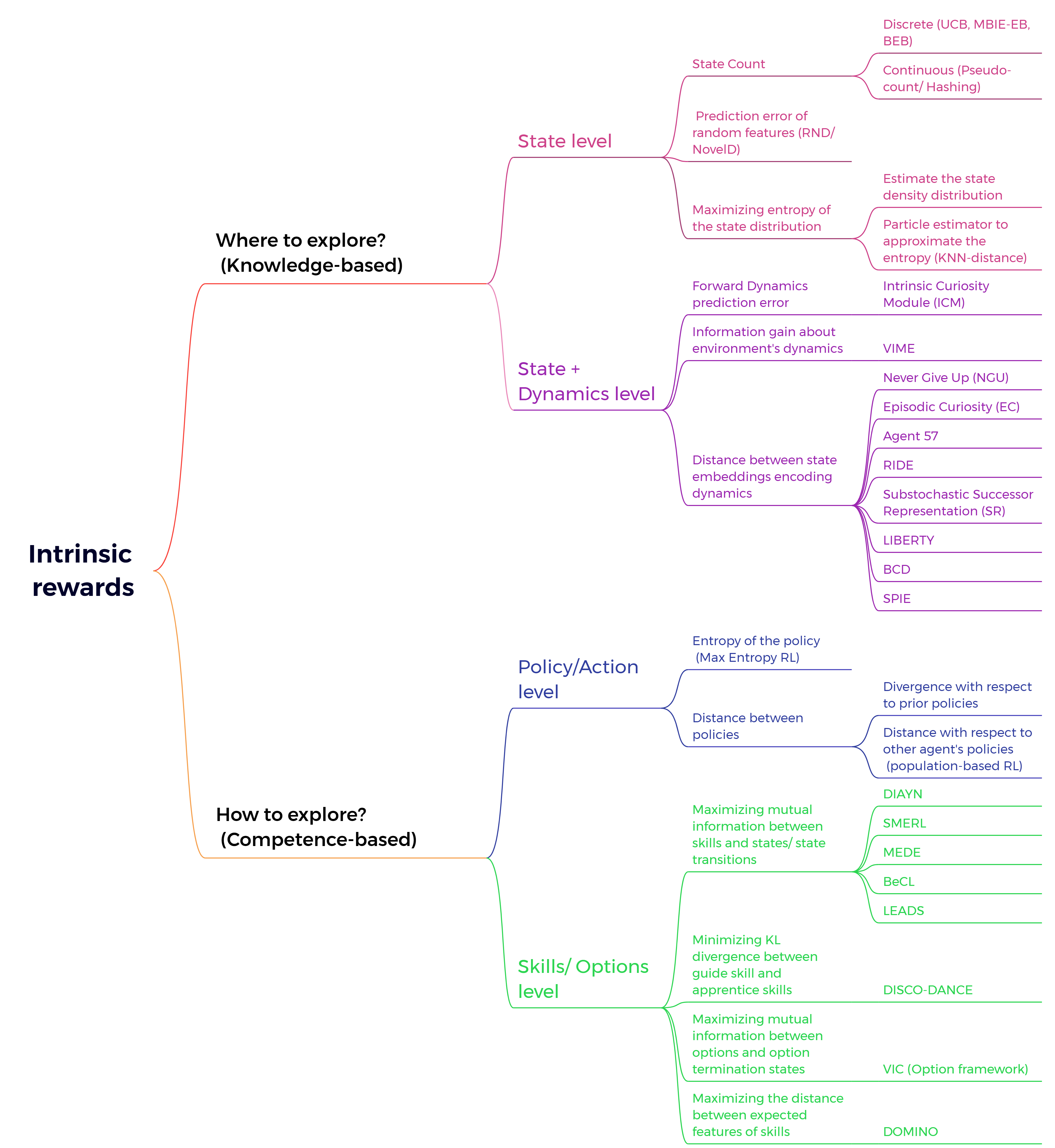}
\caption{Categorisation of the different levels of diversity incurred by intrinsic rewards for exploration in RL.}
\label{diversity map}
\end{figure}

We divide intrinsic rewards into two categories: ``Where to
explore''  and ``How to explore?'', as described in the following and shown in Figure \ref{diversity map}.
\label{sec:appendix1}
\subsection{\textit{\textbf{``Where to explore?''}}}
\label{where to explore}

{\bf State level diversity} \\
%\subsubsection{State level diversity}
%In this subcategory, we collected all the works in which the agent operates on the states without considering the environment's dynamics. The intrinsic reward is a function of the state only: $r_{int}= f(s)$.
In this subcategory, we collect all the works which encourage the exploration of unseen states.
The most common method is ``State Count'', which stores the visitation count of each state, and gives  high intrinsic rewards to encourage revisiting states with low counts \citep{jin2018q, strehl2008analysis,kolter2009near}. While counting works well in tabular cases, it becomes difficult in vast state spaces. Several methods were proposed to extend State Count to large or continuous state spaces, such as pseudo-counts \citep{bellemare2016unifying} and hashing \citep{tang2017exploration}.

Besides count-based methods, features prediction error can be used as a measure of the state novelty. For example, in \citep{burda2018exploration}, authors assessed state novelty by distilling a fixed randomly initialised neural network (target network) into another neural network (predictor network) trained on the data collected by the agent. This technique is called Random Network Distillation (RND), and the main motivation behind it is that the prediction error should be small for frequently visited states. Similarly, the NovelD algorithm \citep{zhang2021noveld} uses RND as a measure of state novelty but it defines the intrinsic reward as the difference in RND prediction errors at two consecutive states $s_t$ and $s_{t+1}$ in a trajectory.

Finally, this level of diversity includes methods which aim to maximise the entropy of the state distribution induced by the policy over finite or infinite horizon by estimating the state density distribution \citep{hazan2019provably, lee2019efficient} or by relying on the K-Nearest Neighbours (KNN) distance as approximation of state entropy \citep{mutti2020policy,liu2021behavior,seo2021state,kim2024accelerating}.

\vspace{1em}
\noindent {\bf State + Dynamics level diversity} \\
This class also aims to visit diverse states but the difference with respect to State level is that the agent considers the novelty of the dynamics as well (not only states) to drive exploration. The agent either tries to build an accurate dynamical model of the environment or learns a dynamics-relevant state representation for exploration.

This subcategory mainly includes curiosity-driven methods which use the forward dynamics prediction error as intrinsic reward, such as \citep{pathak2017curiosity} and \citep{burda2018large}. The key intuition is to encourage the agent to revisit the unfamiliar state transitions where the prediction error is high (high mismatch between the expectation and true experience of the agent). Another curiosity-driven  technique is Variational Information Maximizing Exploration (VIME) \citep{houthooft2016vime}, which pushes the agent to explore states leading to a larger change in the dynamics model (higher information gain).

Moreover, this subcategory includes techniques that estimate the state novelty within a feature space designed to capture the temporal or dynamical aspects of states. For instance, Exploration via Elliptical Episodic Bonuses (E3B) \citep{henaff2022exploration} and RIDE \citep{raileanu2020ride} both utilise an inverse dynamics model (ICM) to learn state embeddings that represent the controllable dynamics of the environment. While RIDE encourages the agent to select actions that produce substantial changes in the state embedding, E3B applies an elliptical episodic bonus to guide exploration. Additional examples include Never Give Up (NGU) \citep{badia2020never}, Agent 57 \citep{badia2020agent57}, and Episodic Curiosity (EC) \citep{savinov2018episodic}, all of which employ memory-based methods using distance-based metrics in a dynamics-aware feature space to approximate State + Dynamics novelty. \review{Similarly, \citep{wang2024efficient} propose LIBERTY approach, which utilises an inverse dynamic bisimulation metric to measure distances between states in a latent space, ensuring effective exploration and policy invariance. The work of \citep{zhu2024abstract} also presents a novel behavioural metric with Cyclic Dynamics (BCD), leveraging successor features and vector quantisation to evaluate behavioural similarity between states and capture interrelations among environmental dynamics. Finally, \citep{machado2020count} propose using the inverse of the norm of the successor representation (SR) as an intrinsic reward to account for transition dynamics. More recently, \citep{yu2024successor} developed the SPIE approach that constructs an intrinsic reward by integrating both prospective and retrospective information from previous trajectories, also based on SR.} %Their method penalises transitions leading to states frequently reached from many other states, encouraging the agent to explore states that are typically more challenging to access. }
%\review{Similarly, \citep{wang2024efficient}  measure the distance between states in a latent space using the bisimulation metric. Specifically, they propose a potential function based on the inverse dynamic bisimulation metric, enabling effective exploration and policy invariance. The work by~\citep{zhu2024abstract} also propose a novel Behavioral metric with Cyclic Dynamics (BCD), utilizing successor features and vector quantization technique to measure the behavioral similarity between states while capturing the interrelations among environmental dynamics.} Finally, authors in \citep{machado2020count} use the inverse of the norm of the successor representation as intrinsic reward, to capture the transition dynamics. \review{A more recent algorithm by~\citep{yu2024successor} constructs an intrinsic reward by incorporating both prospective and retrospective information from previous trajectories. This approach penalises transitions that lead to states frequently reached from numerous other states, and encourages the agent to seek out states that are generally more challenging to reach.}

\subsection{\textit{\textbf{``How to explore?''}}} 

\label{How to explore}
{\bf Policy/Action level diversity} \\
Algorithms in this subcategory aim to explore diverse actions from the same state. %The intrinsic reward is a function of the policy here: $r_{int}=f(\pi(\cdot|s_t))$.
What makes it different from the State + Dynamics algorithms introduced in~\ref{where to explore} is that the previous category uses knowledge about states and dynamics of the environment, and pushes for exploring the areas where the agent knows the least (high uncertainty). In contrast, this level of diversity considers the previous exploration behaviour represented by the policy (how the agent has explored) and pushes it to explore differently, inducing diversity on the policy learned. For example, in Maximum Entropy RL (Max Entropy), the aim is to learn the optimal behaviour while acting as randomly as possible. The objective function becomes the sum of expected rewards and conditional action entropy \citep{eysenbach2021maximum}. Soft-Actor Critic (SAC) \citep{haarnoja2018soft} is a popular RL algorithm implementing the Max Entropy RL framework. Diversity-driven exploration strategy \citep{hong2018diversity} is another exploration technique that encourages the agent to behave differently in similar states. It maximises the divergence between the current policy and prior polices. Similarly, Adversarially Guided Actor-Critic (AGAC) \citep{flet2021adversarially} maximises the divergence between the prediction of the policy and an adversary policy trained to mimic the behavioural policy. The main motivation is to encourage the policy to explore different behaviours by remaining different from the adversary. Another branch which belongs to this diversity level is the population-based exploration, which combines evolutionary strategies with Reinforcement Learning. These approaches train a population of agents to learn diverse behaviours which are high scoring at the same time, in order to effectively explore the environment \citep{conti2018improving,parker2020effective}. \review{For more details on the connection between evolutionary approaches and RL, please refer to the comprehensive survey by~\citep{li2024bridging}.}

\vspace{1em}
\noindent {\bf Skill level diversity}\\
Skill level diversity disentangles diverse behaviours into different latent-conditioned policies (also called skills). The policy $\pi$ is conditioned on a latent variable $z \sim p(z)$, and each $z$ defines a different policy denoted by $\pi(a|s,z)$ \citep{cohen2019maximum}. This category aims to discover diverse skills and the intrinsic reward is a function of the skill. Most methods falling in this category come from the domain of unsupervised skill discovery and use a discriminator-based architecture such as Diversity is all you need (DIAYN) \citep{eysenbach2018diversity}. DIAYN replaces the task reward with a learned discriminator term $q_\alpha (z|s)$ that infers the behaviour from the current state, in order to generate diverse policies visiting different set of states. It also uses Max Entropy RL framework to learn skills which are as random as possible \citep{eysenbach2018diversity}. Maximum entropy diverse exploration (MEDE) \citep{cohen2019maximum} is very similar to ``DIAYN + extrinsic reward'', with the small difference of conditioning the discriminator on the state-action pair $q_\alpha (z|s,a)$ instead of the state only. Moreover, MEDE uses the discriminator term as prior in the objective function instead of adding it as intrinsic reward. Structured Max Entropy RL (SMERL) is another algorithm with the same approach as DIAYN, but it adds the intrinsic reward to the task reward, only when the policies have achieved at least near-optimal return \citep{kumar2020one}. DOMINO also learns diverse policies while remaining near optimal; it uses an intrinsic reward that maximises the diversity of policies by measuring the distance between the expected features of the policies' state-action occupancies \citep{zahavy2022discovering}. It's important to mention that skills in the literature can be called options or goals. For example, Variational intrinsic control (VIC) is an algorithm which provides the agent with an intrinsic reward that relies on modelling options and learning policies conditioned on these options \citep{gregor2016variational}. Instead of sampling options from a fix prior distribution as in DIAYN, VIC learns the prior distribution of options and updates it in order to choose options with higher rewards \citep{gregor2016variational}. Both DIAYN and VIC are part of goal-conditioned RL methods, where goals are internally generated by agents and achieved via self-generated rewards \citep{colas2022autotelic}. \review{More Recent unsupervised skill learning methods have emerged, such as ~\citep{yang2023behavior} which proposed behaviour Contrastive Learning (BeCL), a novel competence-based method that uses contrastive learning to encourage similar behaviours within the same skill and diverse behaviours across different skills. This is done by maximising the mutual information between different states generated by the same skill as an intrinsic reward. Another recent work by~\citet{kim2024learning}
proposed skill discovery with guidance (DISCO-DANCE) which identifies the guide skill most likely to reach unexplored states, directs other skills to follow it, and disperses them to maximise distinctiveness. Moreover, ~\citet{tolguenec2024exploration} proposed LEADS, which maximises a variant of the mutual information between skills and states, by leveraging the successor state measure to tailor skills toward less-visited states while also maximising the disparity between skills.}
%~\citet{celik2024acquiring} using Mixture of Experts, where
%each expert formalises a skill as a contextual motion primitive, optimises each expert
%and its associate context distribution to a maximum entropy objective that incentivises learning diverse skills in similar contexts. 

%through contrastive learning among behaviors,
%which makes the agent produce similar behaviors
%for the same skill and diverse behaviors for different skills. They use the MI between different states generated by the same skill as an
%intrinsic rewards and train a policy to maximise it. Intuitively, BeCL encourages the agent to perform similarly
%condition on the same skill, and performs diversely among
%different skills}

\section{MiniGrid Environments}
\label{sec:appendix2}
\begin{figure}[h]
    \centering
     \begin{subfigure}[b]{0.3\linewidth}
        \includegraphics[width=\linewidth]{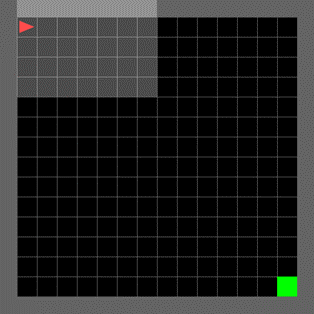} % Change this to your fourth image file path
        \caption{Empty 16x16}
        \label{Empty}
    \end{subfigure}   
    \hfill
    \begin{subfigure}[b]{0.3\linewidth}
        \includegraphics[width=\linewidth]{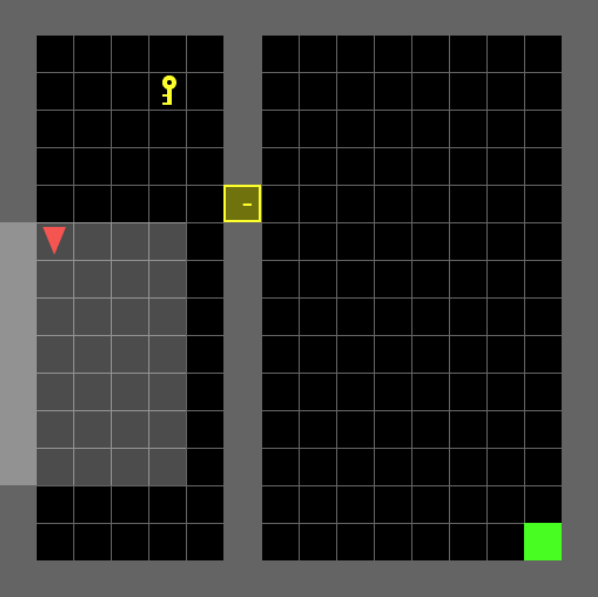}
        \caption{DoorKey 16x16}
        \label{Doorkey}
    \end{subfigure}
    \hfill
    \begin{subfigure}[b]{0.25\linewidth}
    \includegraphics[width=\linewidth]{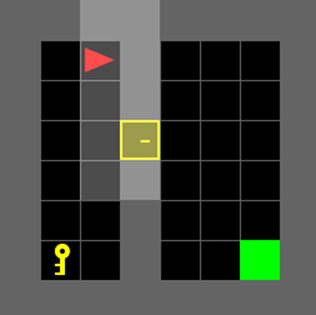}
    \caption{DoorKey 8x8}
    \label{Doorkey8}
  \end{subfigure}

    \vspace{0.5cm}
    \begin{subfigure}[b]{0.5\linewidth}
        \includegraphics[width=\linewidth]{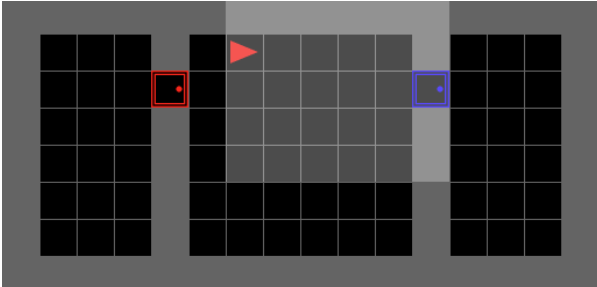}
        \caption{RedBlueDoors 8x8}
        \label{RedblueDoors}
    \end{subfigure}  
    \hfill
    \begin{subfigure}[b]{0.3\linewidth}
        \includegraphics[width=\linewidth]{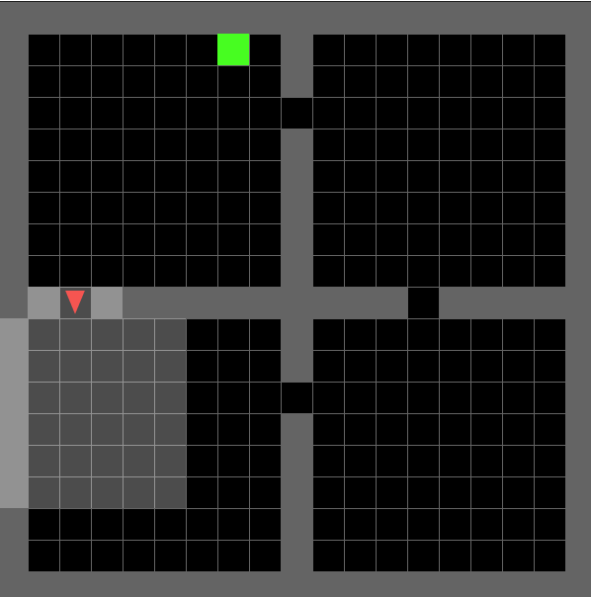}
        \caption{FourRooms}
        \label{FourRooms}
    \end{subfigure}
  \vspace{0.5cm}
    \caption{MiniGrid environments}
    \label{fig:overall}
\end{figure}
\noindent We use the following MiniGrid environments shown in Figure \ref{fig:overall}:
\begin{enumerate}

\item Empty: This is an empty grid, where the agent is always placed in the corner opposite to the goal. The task is to get to the green goal square. We use the regular variant ``MiniGrid-Empty-16x16-v0''.
\item DoorKey: This is a sparse reward environment which requires a certain order of visiting the states to solve the task; the agent needs to pick up the key, open the door then get to the green goal square. It does not get any reward after picking up the key or unlocking the door; it gets rewarded just at the end of the task. We use ``MiniGrid-DoorKey-16x16-v0'' in the case of grid encodings and ``MiniGrid-DoorKey-8x8-v0'' in the case of RGB observations.
\item FourRooms: In this environment, the agent must navigate a maze consisting of four rooms, with both its initial position and goal position being randomised. We use ``MiniGrid-FourRooms-v0'' where each of the four rooms consists of a grid of size $8 \times 8$.
\item RedBlueDoors : The agent is randomly placed in a room where there are one red and one blue door facing opposite directions. The task consists of opening the red door before opening the blue door. The agent must rely on its memory of whether it has previously opened the other door to successfully complete the task, as it cannot see the door behind it. We use ``MiniGrid-RedBlueDoors-8x8-v0''.

\end{enumerate}

For all tasks, a maximum number of steps $t_{max}$ is assigned, to encourage the agent to solve the task as fast as possible. When the agent succeeds after $t$ steps, it gets a reward $r=1-0.9*t/t_{max}$ in all three environments. The episode ends when the agent collects the final reward or when the maximum number of steps is exceeded.
\paragraph{Observation and action spaces:}
The observations are egocentric and partially observable. We considered first the grid encoding observations of size $7 \times 7 \times 3$. The first two dimensions ($7\times7$) compose the tile set, and the last dimension encode the object type (wall, door, $\cdots$), the object colour (red, green, $\cdots$) and the object status (door open, door closed, door locked). Specifically, object type $ \in \{0,1,2,3,4,5,6,7,8,9,10\}$, object colour $\in \{0,1,2,3,4,5\}$, and object status $\in \{0,1,2\}$.
Then, we used partial RGB visual observations of size $56 \times 56 \times 3$ ($7$ tiles of $8 \times 8$ pixels each) to increase the complexity of the task since agents must extract features directly from the images.
There are 7 actions available to the agent:  turn left, turn right, move forward,
pick up an object, drop an object, toggle and done. Some of these actions are unused in certain tasks.

\section{Hyperparameters}
\label{sec:appendix4}
 For State Count, and ICM, we use the hyperparameters of the previous study \citep{andres2022evaluation}. Since Max Entropy + PPO and DIAYN were not tested before on MiniGrid, we run a grid search over $\beta \in [0.1,0.01,0.001,0.0005]$ and pick the best values of $\beta$ which result in the highest return during training. The chosen values of $\beta$ are summarised in Table \ref{intrinsic reward coefficient}. For DIAYN, we choose to train 10 skills, which is the number used in the study by \citep{gaya2021learning}, and we use a discriminator learning rate of $3 \times e^{-4}$ following the implementation of DIAYN paper \citep{eysenbach2018diversity} (Table \ref{PPO hyperparams}). Note that we reused the same hyperparameters for the second part where we tested on RGB observations.
 \vspace{4cm}

\begin{table}[h]
   \caption{List of hyperparameters}
  \centering
  \begin{tabular}{lll}
    \toprule  
    Number of parallel actors     & 16  \\
    Number of frames per rollout & 128    \\
    Number of epochs     & 4     \\
    Batch size      & 256     \\
    Discount $\gamma$     &  0.99     \\
    Learning rate     & 0.0001      \\
    GAE $\lambda$     & 0.95       \\
    Entropy regularisation coefficient     & 0.0005  \\
    Value loss coefficient     & 0.5     \\
    Clipping factor PPO     & 0.2  \\
    Gradient clipping    & 0.5     \\
    \midrule
    Forward dynamics loss coefficient    & 10  \\
    Inverse dynamics loss coefficient    & 0.1 \\
    Learning rates (state embedding, forward, and inverse dynamics)    & 0.0001\\
    Number of skills & 10 \\
    Discriminator learning rate & 0.0003 \\
    SimHash key size K & 16 \\
    \bottomrule
  \end{tabular}

   \label{PPO hyperparams}
\end{table}
\begin{table}[h]
  \caption{Best intrinsic reward coefficients $\beta$}
  \centering
  \begin{tabular}{l|l|l|l|l|}
    \hline
    \multicolumn{1}{|c|}{} 
    & Empty & DoorKey & RedBlueDoors & FourRooms \\ \hline
    \multicolumn{1}{|l|}{State Count} & 0.005 & 0.005 & 0.005 & 0.005 \\ \hline
    \multicolumn{1}{|l|}{Max Entropy} & 0.0005 & 0.0005 & 0.0005 &0.0005 \\ \hline
    \multicolumn{1}{|l|}{ICM} & 0.05 & 0.05 & 0.05 & 0.05 \\ \hline
    \multicolumn{1}{|l|}{DIAYN} & 0.01 & 0.01 &0.01 &0.01 \\ \hline
  \end{tabular}
  \label{intrinsic reward coefficient}
\end{table}
\vspace{2cm}
\section{Additional experimental results}
\label{sec:appendix5}
\subsection{Grid Encoding Observation space}  \label{sec:appendix_NonRGB_results}

\begin{table}[h]
\caption{Frame number at which the reward is found for the first, second, and third time by each exploration method on Empty 16x16 environment with grid encodings observations. Results are averaged over five runs. Mean and standard deviation ($\mu \pm \sigma$) are reported.}
\centering
\begin{tabular}{|c|l|l|l|}
\hline
Empty 16x16           & \multicolumn{1}{c|}{First reward} & \multicolumn{1}{c|}{Second Reward} & \multicolumn{1}{c|}{Third reward} \\ \hline
PPO               & 15 452   $\pm$ 7112           & 21 273   $\pm$ 11539               & 28 304  $\pm$ 14785        \\ \hline
PPO + State Count & 18 428   $\pm$ 9119           & 25 340   $\pm$ 11537             & 32 483  $\pm$ 12888   \\ \hline
PPO + Max Entropy     & 16 841   $\pm$ 6916           & 22 768 $\pm$ 6736                & 27 318  $\pm$ 10355         \\ \hline
PPO + ICM         & \textbf{8 918  $\pm$ 3565}          & \textbf{13 436  $\pm$ 6830} & \textbf{18 281  $\pm$ 9467}           \\ \hline
PPO + DIAYN pretraining      & 12 668  $\pm$ 7030          & 20 076  $\pm$ 13898                        & 326 963  $\pm$ 662300 \\ \hline
PPO + DIAYN finetuning      & 1 001 862   $\pm$ 1187338  & 1 130 208   $\pm$ 1082785   & 1 207 600  $\pm$ 1038924 \\ \hline
\end{tabular}
\label{table_empty_nonRGB}
\end{table}

\begin{table}[h]
\caption{Frame number at which the reward is found for the first, second, and third time by each exploration method on DoorKey 16x16 environment with grid encodings observations. Results are averaged over five runs. Mean and standard deviation ($\mu \pm \sigma$) are reported. If the reward is never found, the frame number is set to the training budget (40M).}
\centering
\begin{tabular}{|c|l|l|l|}
\hline
DoorKey 16x16          & \multicolumn{1}{c|}{First reward} & \multicolumn{1}{c|}{Second Reward} & \multicolumn{1}{c|}{Third reward} \\ \hline
PPO               & 1 242 342  $\pm$ 529863           & 2 276 508  $\pm$ 917486             & 3 186 537  $\pm$ 1616226        \\ \hline
PPO + State Count & \textbf{496 486  $\pm$ 550012}    & \textbf{558 204  $\pm$ 548684}    & \textbf{783 075  $\pm$ 615917}   \\ \hline
PPO + Max Entropy     & 594 649  $\pm$ 696956           & 1 067 401 $\pm$ 743704                & 3 300 668 $\pm$ 3108002          \\ \hline
PPO + ICM         & 1 089 286 $\pm$ 734419           & 1 287 632 $\pm$ 674758              & 1 683 612 $\pm$ 539173           \\ \hline
PPO + DIAYN Pretraining      & 40 000 000 $\pm$ 0           & 40 000 000 $\pm$ 0           & 40 000 000 $\pm$ 0 \\ \hline
PPO + DIAYN finetuning      & 2 087 398  $\pm$ 449537 & 2 221 756  $\pm$ 447746   & 2 516 739  $\pm$ 689340 \\ \hline
\end{tabular}
\label{table_doorkey}
\end{table}

\begin{table}[h]
\caption{Frame number at which the reward is found for the first, second, and third time by each exploration method on RedBlueDoors environment with grid encodings observations. Results are averaged over five runs. Mean and standard deviation ($\mu \pm \sigma$) are reported.}
\centering
\begin{tabular}{|c|l|l|l|}
\hline
RedBlueDoors          & \multicolumn{1}{c|}{First reward} & \multicolumn{1}{c|}{Second Reward} & \multicolumn{1}{c|}{Third reward} \\ \hline
PPO               & 13 136   $\pm$ 5647           & \textbf{17 568   $\pm$ 8303}               & 26 553  $\pm$ 6733        \\ \hline
PPO + State Count & 13 180   $\pm$ 8236           & 25 923   $\pm$ 11537             & 33 545  $\pm$ 19115   \\ \hline
PPO + Max Entropy     & \textbf{9 417   $\pm$ 2678}          & 20 464 $\pm$ 10420                & \textbf{24 432  $\pm$ 10339}         \\ \hline
PPO + ICM         & 37 721  $\pm$ 68636          & 129 043  $\pm$ 175 507           & 162 060  $\pm$ 193005           \\ \hline
PPO + DIAYN pretraining      & 19 244  $\pm$ 10004          & 24 611  $\pm$ 12661                        & 39 280  $\pm$ 25334 \\ \hline
PPO + DIAYN finetuning      & 2 992 614   $\pm$ 2118551  & 3 006 659  $\pm$ 2114575   & 3 033 043  $\pm$ 2096962 \\ \hline
\end{tabular}
\label{table_redbluedoors_nonRGB}
\end{table}

\begin{table}[h]
\caption{Frame number at which the reward is found for the first, second, and third time by each exploration method on FourRooms environment with grid encodings observations. Results are averaged over five runs. Mean and standard deviation ($\mu \pm \sigma$) are reported.}
\centering
\begin{tabular}{|c|l|l|l|}
\hline
FourRooms          & \multicolumn{1}{c|}{First reward} & \multicolumn{1}{c|}{Second Reward} & \multicolumn{1}{c|}{Third reward} \\ \hline
PPO               & 25 222   $\pm$ 32606           & 97 033   $\pm$ 41446               & 150 188  $\pm$ 104821        \\ \hline
PPO + State Count & 15 465   $\pm$ 9712           & 34 649   $\pm$ 11090             & 51 820  $\pm$ 23054   \\ \hline
PPO + Max Entropy     &  2 479 424  $\pm$ 5498212          & 5 327 913 $\pm$ 5306632                 & 6 874 905  $\pm$ 5056693         \\ \hline
PPO + ICM         & 89 433  $\pm$ 111832        & 197 312  $\pm$ 171435           & 274 883  $\pm$ 171782           \\ \hline
PPO + DIAYN pretraining      & 2 089  $\pm$ 1178          & 9 049  $\pm$ 5350                        & 14 531  $\pm$ 9099 \\ \hline
PPO + DIAYN finetuning      & 29 238   $\pm$ 27103  & 41 376  $\pm$ 35912   & 69 737  $\pm$ 53656 \\ \hline
\end{tabular}
\label{table_fourRooms_nonRGB}
\end{table}

\begin{figure}[h]
    % Top right: Empty environment image
    \hspace*{-0.37cm} % Negative space to shift left
    \makebox[0.95\textwidth][r]{ % Move content to the right edge of the line
        \begin{minipage}[t]{0.13\linewidth}
            \caption*{Empty}
            \includegraphics[width=\linewidth]{Empty_env.png} % Empty environment image file
           
        \end{minipage}
    }%
    
    % Bottom: Full-width heatmap
    \begin{minipage}[t]{\linewidth}
        \centering
        \includegraphics[width=\linewidth]{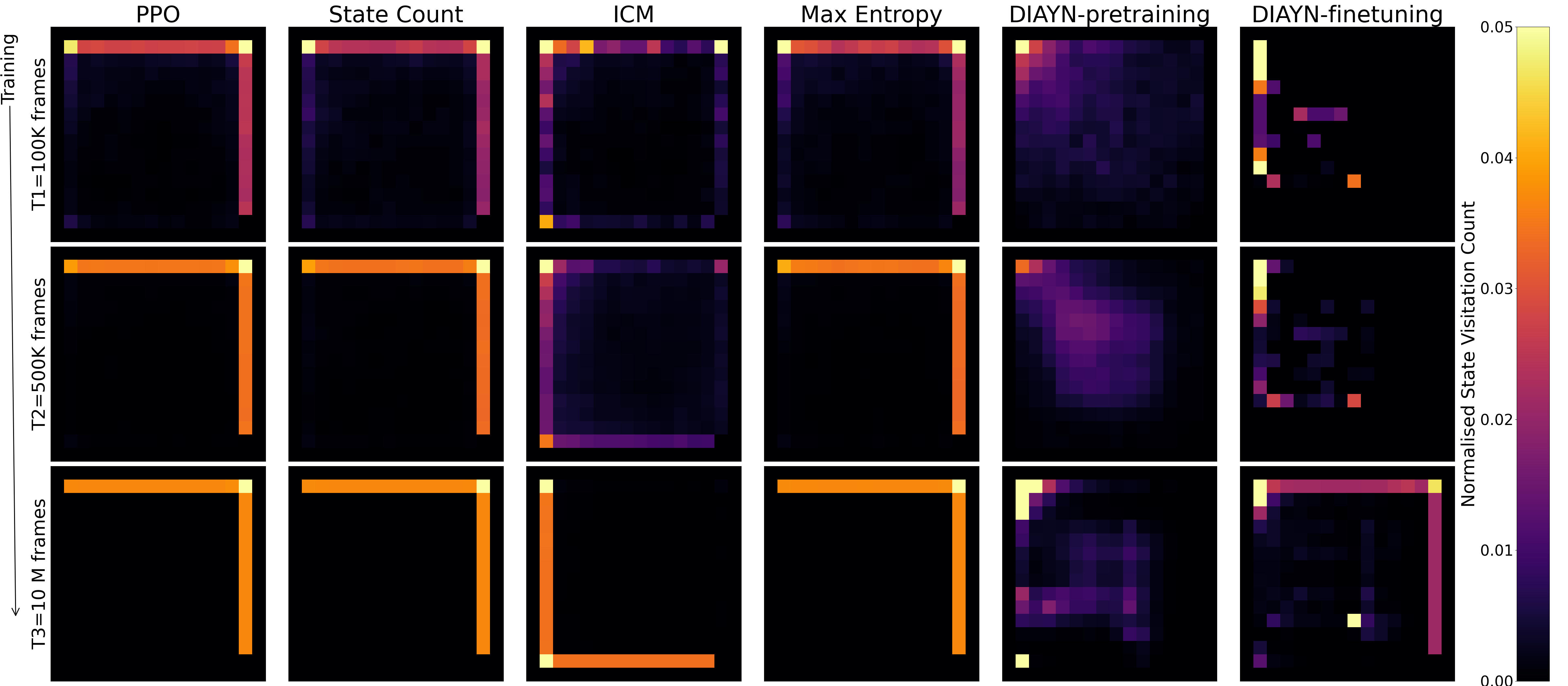} % Heatmap image file
        \caption{State visitation count during training for 10M frames on singleton Empty 16x16 environment with grid encodings observations. For each intrinsic reward method, snapshots of the heatmap are taken at three different timesteps T1: 100K frames, T2: 500K frames and T3: 10M frames. Colour intensity represents the proportion of frames spent in each state, with high values capped for better visualisation.}
        \label{heatmap_Empty_NonRGB}
    \end{minipage}
\end{figure}

\begin{figure}[h]
\hspace*{-0.37cm} % Negative space to shift left
    \makebox[0.95\textwidth][r]{ % Move content to the right edge of the line
        \begin{minipage}[t]{0.13\linewidth}
            \caption*{DoorKey 16x16}
            \includegraphics[width=\linewidth]{DoorKey_env.png} % 
          
        \end{minipage}
    }%
    
    % Bottom: Full-width heatmap
    \begin{minipage}[t]{\linewidth}
        \centering
        \includegraphics[width=\linewidth]{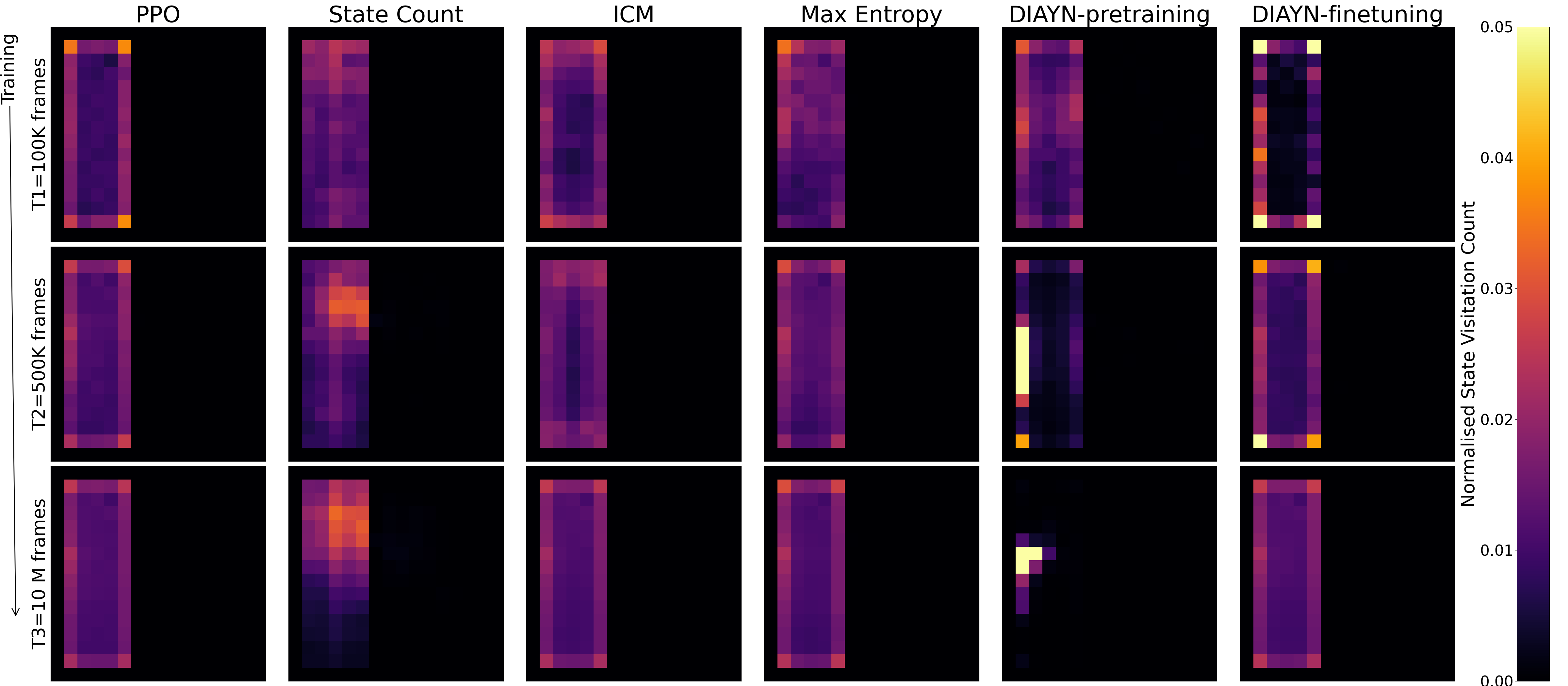} % Heatmap image file
        \caption{State visitation count during training for 10M frames on singleton DoorKey 16x16 environment with grid encodings observations. For each intrinsic reward method, snapshots of the heatmap are taken at three different timesteps T1: 100K frames, T2: 500K frames and T3: 10M frames. Colour intensity represents the proportion of frames spent in each state, with high values capped for better visualisation.}
        \label{heatmap_DoorKey_NonRGB}
    \end{minipage}
\end{figure}

\begin{figure}[h]
\hspace*{-0.37cm} % Negative space to shift left
    \makebox[0.95\textwidth][r]{ % Move content to the right edge of the line
        \begin{minipage}[t]{0.13\linewidth}
            \caption*{FourRooms}
            \includegraphics[width=\linewidth]{FourRooms_env.png} % 
        \end{minipage}
    }%
    
    % Bottom: Full-width heatmap
    \begin{minipage}[t]{\linewidth}
        \centering
        \includegraphics[width=\linewidth]{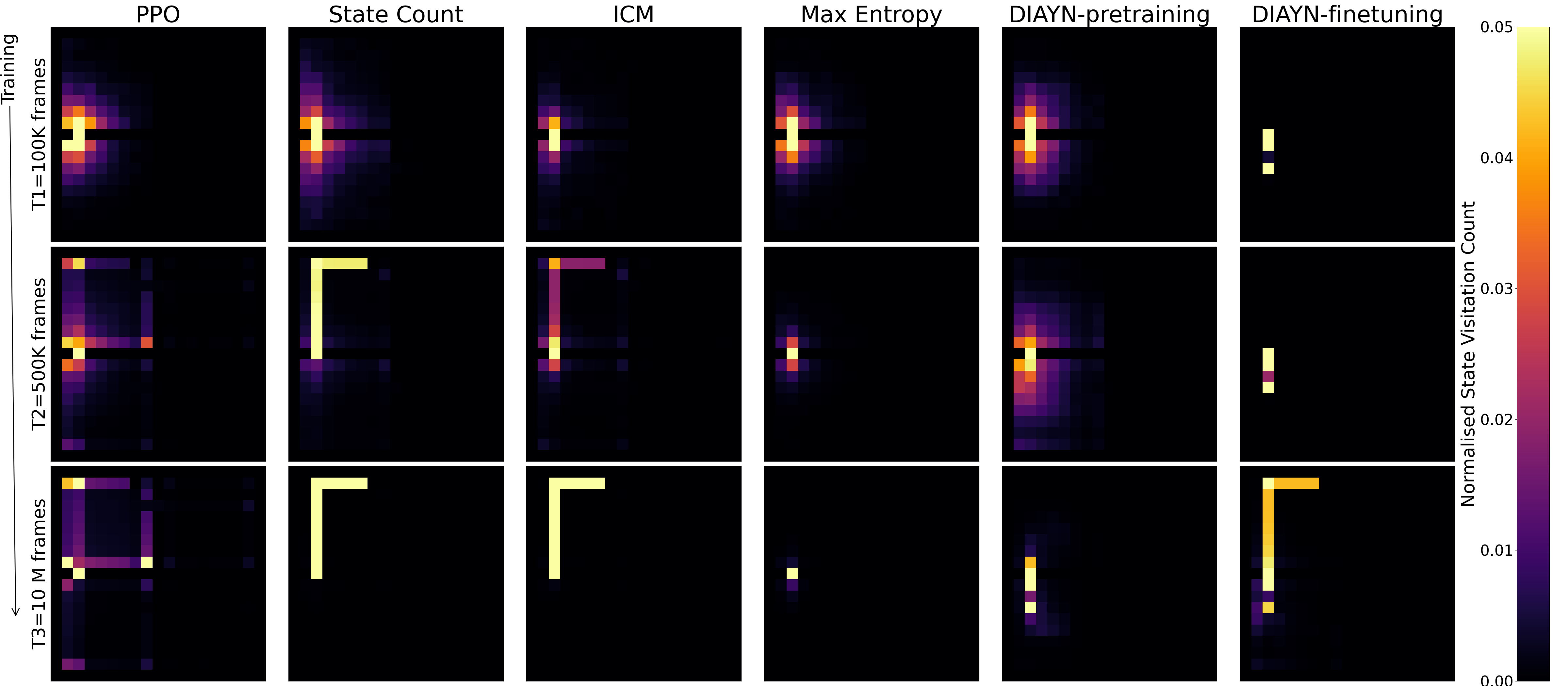} % Heatmap image file
        \caption{State visitation count during training for 10M frames on singleton FourRooms environment with grid encodings observations. For each intrinsic reward method, snapshots of the heatmap are taken at three different timesteps T1: 100K frames, T2: 500K frames and T3: 10M frames. Colour intensity represents the proportion of frames spent in each state, with high values capped for better visualisation.}
        \label{FouRooms_NonRGB}
    \end{minipage}
\end{figure}

\begin{figure}[h]
\hspace*{-0.3cm} % Negative space to shift left
    \makebox[0.95\textwidth][r]{ % Move content to the right edge of the line
        \begin{minipage}[t]{0.15\linewidth}
            \caption*{RedBlueDoors}
            \includegraphics[width=\linewidth]{RedBlue_env.png} % 
        \end{minipage}
    }%
    
    % Bottom: Full-width heatmap
    \begin{minipage}[t]{\linewidth}
        \centering
        \includegraphics[width=\linewidth]{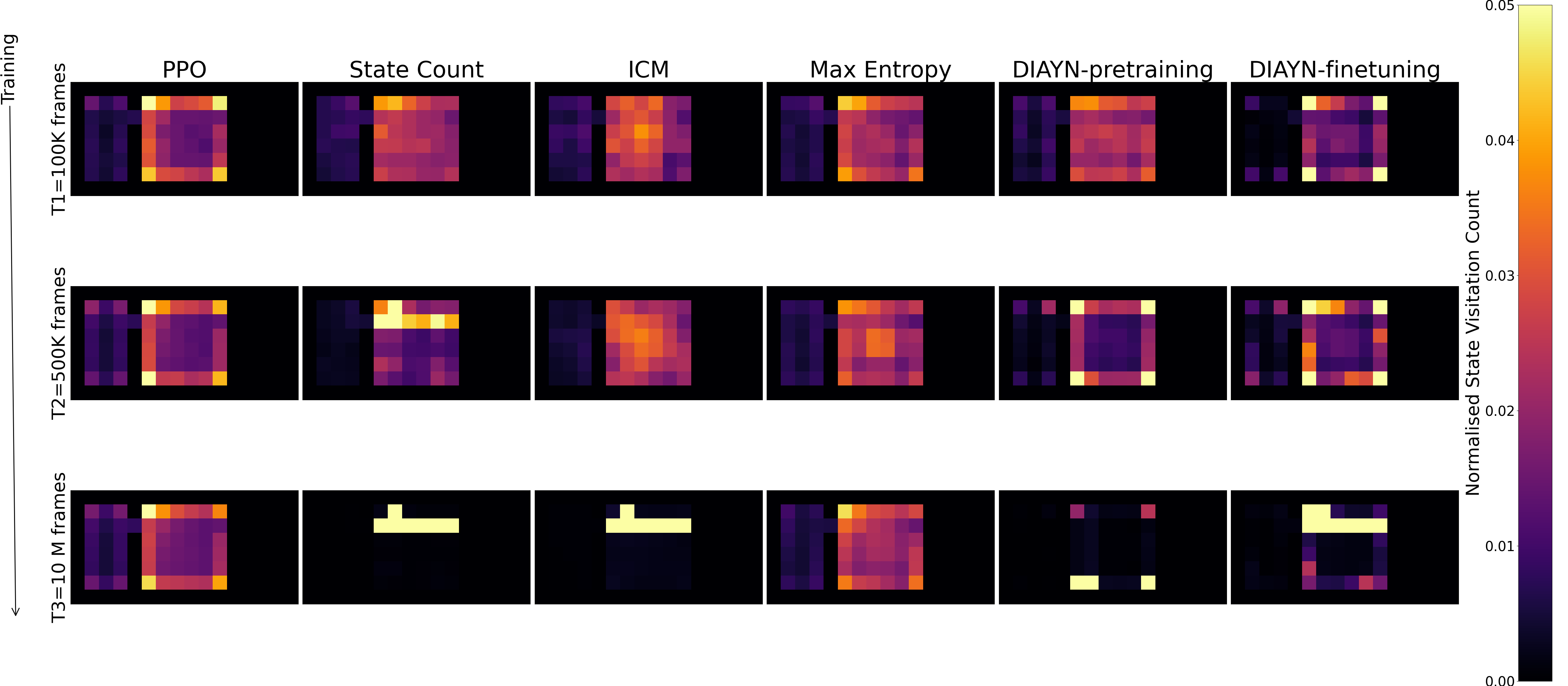} % Heatmap image file
        \caption{State visitation count during training for 10M frames on singleton RedBlueDoors environment with grid encodings observations. For each intrinsic reward method, snapshots of the heatmap are taken at three different timesteps T1: 100K frames, T2: 500K frames and T3: 10M frames. Colour intensity represents the proportion of frames spent in each state, with high values capped for better visualisation.}
        \label{RedBlueDoors_NonRGB}
    \end{minipage}
\end{figure}

\FloatBarrier
\subsection{RGB Observation space} \label{sec:appendix_RGB_results}

\begin{table}[h]
\caption{Frame number at which the reward is found for the first, second, and third time by each exploration method on Empty 16x16 environment with RGB observations. Results are averaged over five runs. Mean and standard deviation ($\mu \pm \sigma$) are reported. }
\centering
\begin{tabular}{|c|l|l|l|}
\hline
Empty 16x16 RGB              & \multicolumn{1}{c|}{First reward} & \multicolumn{1}{c|}{Second Reward} & \multicolumn{1}{c|}{Third reward} \\ \hline
PPO               & 48 256   $\pm$ 85183           & 103 401    $\pm$ 112583               & 115 148   $\pm$ 117261        \\ \hline
PPO + SimHash    & 71 132    $\pm$ 9119           & 75 686    $\pm$ 95328             & 79 699  $\pm$ 96431   \\ \hline
PPO + Max Entropy     & \textbf{43 072   $\pm$ 76270}  & \textbf{49 033  $\pm$ 79868}     & \textbf{59 494   $\pm$ 81361}         \\ \hline
PPO + ICM         & 448 121   $\pm$ 500641          & 493 401   $\pm$ 522557          & 509 750   $\pm$ 509223           \\ \hline
PPO + DIAYN pretraining      & 896 963   $\pm$ 1257009         & 2 262 649   $\pm$ 3635427                        & 2 267 164   $\pm$ 3639831 \\ \hline
PPO + DIAYN finetuning      & 69 712   $\pm$ 102945  & 94 240   $\pm$ 138306   & 110 710  $\pm$ 135551 \\ \hline
\end{tabular}
\label{table_empty_RGB}
\end{table}

\begin{table}[h]
\caption{Frame number at which the reward is found for the first, second, and third time by each exploration method on DoorKey 8x8 environment with RGB observations. Results are averaged over five runs. Mean and standard deviation ($\mu \pm \sigma$) are reported. If the reward is never found, the frame number is set to the training budget (40M).}
\centering
\begin{tabular}{|c|l|l|l|}
\hline
DoorKey 8x8 RGB              & \multicolumn{1}{c|}{First reward} & \multicolumn{1}{c|}{Second Reward} & \multicolumn{1}{c|}{Third reward} \\ \hline
PPO               & 31 430    $\pm$ 39987           & \textbf{49 257     $\pm$ 44984}              & \textbf{75 587    $\pm$ 32508}        \\ \hline
PPO + SimHash    & 93 494     $\pm$ 75207           & 115 529    $\pm$ 80404             & 143 840   $\pm$ 71717   \\ \hline
PPO + Max Entropy     & \textbf{26 870    $\pm$ 34213}         & 100 931   $\pm$ 96891               & 124 828    $\pm$ 114469        \\ \hline
PPO + ICM         &   445 222    $\pm$ 433991          & 655 200    $\pm$ 384520      & 713 795    $\pm$ 381101          \\ \hline
PPO + DIAYN pretraining      & 32 003 513    $\pm$ 17880687       & 40 000 000    $\pm$ 0                        & 40 000 000   $\pm$ 0 \\ \hline
PPO + DIAYN finetuning      & 40 000 000    $\pm$ 0     & 40 000 000    $\pm$ 0  & 40 000 000   $\pm$ 0 \\ \hline
\end{tabular}
\label{table_Doorkey_RGB}
\end{table}

\begin{table}[h]
\caption{Frame number at which the reward is found for the first, second, and third time by each exploration method on RedBlueDoors environment with RGB observations. Results are averaged over five runs. Mean and standard deviation ($\mu \pm \sigma$) are reported. If the reward is never found, the frame number is set to the training budget (40M).}
\centering
\begin{tabular}{|c|l|l|l|}
\hline
RedBlueDoors RGB              & \multicolumn{1}{c|}{First reward} & \multicolumn{1}{c|}{Second Reward} & \multicolumn{1}{c|}{Third reward} \\ \hline
PPO               & 18 504     $\pm$ 12321           & \textbf{28 179      $\pm$ 19650 }              & \textbf{44 342     $\pm$ 36719}        \\ \hline
PPO + SimHash    & 35 516      $\pm$ 38700           & 49 548     $\pm$ 48897            & 60 643   $\pm$ 58630   \\ \hline
PPO + Max Entropy     & 51 871     $\pm$ 51587         & 71 776    $\pm$ 59120               & 97 907     $\pm$ 88967       \\ \hline
PPO + ICM         &   \textbf{18 355     $\pm$ 26236}          & 206 547     $\pm$ 420774     & 219 718    $\pm$ 419780          \\ \hline
PPO + DIAYN pretraining      & 16 012 892   $\pm$ 21897134       & 16 015 049    $\pm$ 21895167                        & 24 011 241   $\pm$ 21893513 \\ \hline
PPO + DIAYN finetuning      & 24 212 716   $\pm$ 21618947     & 24 212 716    $\pm$ 21618947  & 24 219 180   $\pm$ 21610106 \\ \hline
\end{tabular}
\label{table_RedBlueDoors_RGB}
\end{table}

\begin{table}[h]
\caption{Frame number at which the reward is found for the first, second, and third time by each exploration method on FourRooms environment with RGB observations. Results are averaged over five runs. Mean and standard deviation ($\mu \pm \sigma$) are reported.}
\centering
\begin{tabular}{|c|l|l|l|}
\hline
FourRooms RGB              & \multicolumn{1}{c|}{First reward} & \multicolumn{1}{c|}{Second Reward} & \multicolumn{1}{c|}{Third reward} \\ \hline
PPO               & 6 057      $\pm$ 7369           & 27 203       $\pm$ 34679              & 41 766     $\pm$ 47238        \\ \hline
PPO + SimHash    & 7 561       $\pm$ 13820          & 13 171     $\pm$ 11599          & 20 406    $\pm$ 17137   \\ \hline
PPO + Max Entropy     & 7 654      $\pm$ 13591       & 9 014     $\pm$ 13184               & 13 907     $\pm$ 12435       \\ \hline
PPO + ICM         & 7 491     $\pm$ 10928          & 10 060     $\pm$ 12737     & 16 912    $\pm$ 16744        \\ \hline
PPO + DIAYN pretraining      & 3 276 505    $\pm$ 7322741       & 3 290 252     $\pm$ 7342741                        & 3 296 742   $\pm$ 7343739 \\ \hline
PPO + DIAYN finetuning      & \textbf{4 470   $\pm$ 2384}     & \textbf{6 854     $\pm$ 4956}  & \textbf{8 166    $\pm$ 4735} \\ \hline
\end{tabular}
\label{table_FourRooms_RGB}
\end{table}

\begin{figure}[h]
    % Top right: Empty environment image
    \hspace*{-0.37cm} % Negative space to shift left
    \makebox[0.95\textwidth][r]{ % Move content to the right edge of the line
        \begin{minipage}[t]{0.13\linewidth}
            \caption*{Empty}
            \includegraphics[width=\linewidth]{Empty_env.png} % Empty environment image file
           
        \end{minipage}
    }%
    
    % Bottom: Full-width heatmap
    \begin{minipage}[t]{\linewidth}
        \centering
        \includegraphics[width=\linewidth]{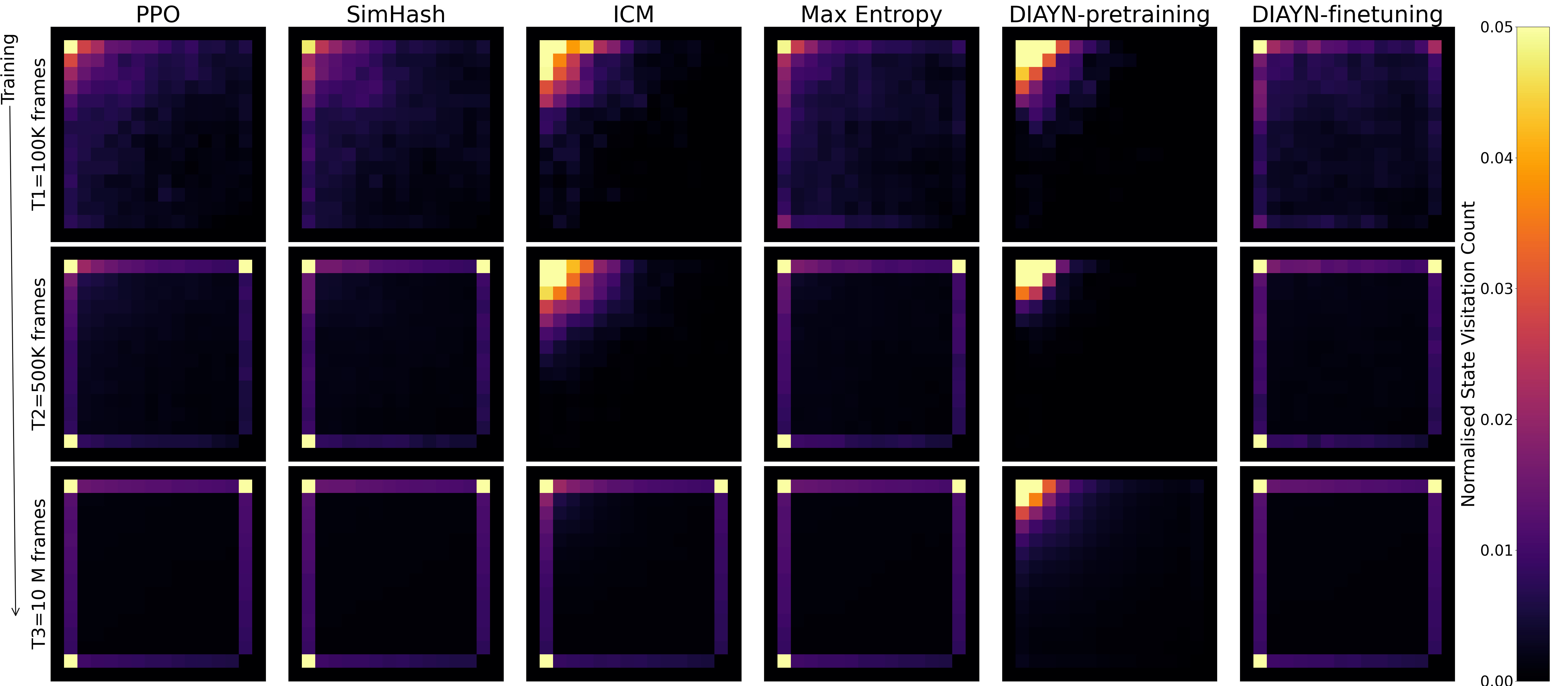} % Heatmap image file
        \caption{State visitation count during training for 10M frames on singleton Empty 16x16 environment with RGB observations. For each intrinsic reward method, snapshots of the heatmap are taken at three different timesteps T1: 100K frames, T2: 500K frames and T3: 10M frames. Colour intensity represents the proportion of frames spent in each state, with high values capped for better visualisation.}
        \label{heatmaps_Empty_RGB}
    \end{minipage}
\end{figure}

\begin{figure}[h]
 \hspace*{-0.37cm} % Negative space to shift left
    \makebox[0.95\textwidth][r]{ % Move content to the right edge of the line
        \begin{minipage}[t]{0.13\linewidth}
            \caption*{DoorKey 8x8}
            \includegraphics[width=\linewidth]{DoorKey8_env.png} % 
          
        \end{minipage}
    }%
    
    % Bottom: Full-width heatmap
    \begin{minipage}[t]{\linewidth}
        \centering
        \includegraphics[width=\linewidth]{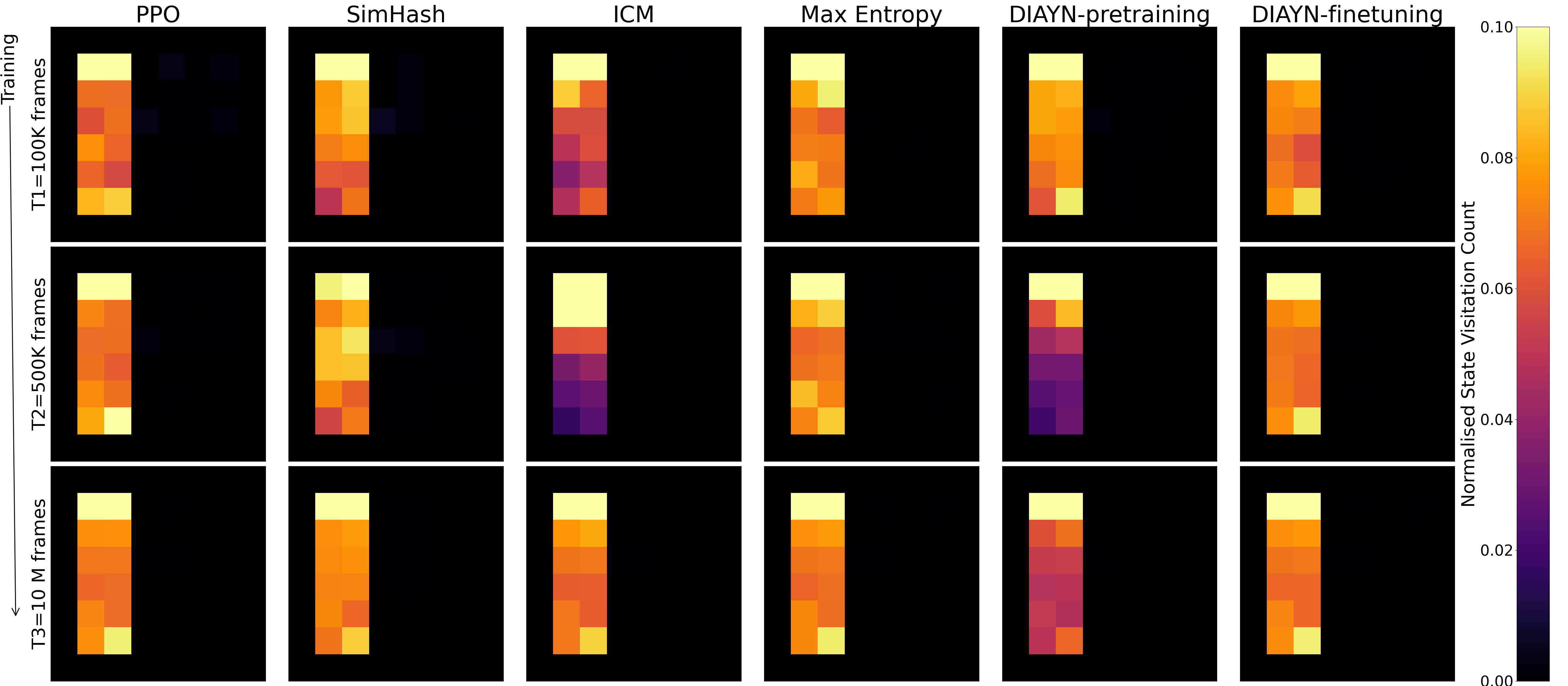} % Heatmap image file
       \caption{State visitation count during training for 10M frames on singleton DoorKey 8x8 environment with RGB observations. For each intrinsic reward method, snapshots of the heatmap are taken at three different timesteps T1: 100K frames, T2: 500K frames and T3: 10M frames. Colour intensity represents the proportion of frames spent in each state, with high values capped for better visualisation.}
        \label{heatmaps_DoorKey_RBG}
    \end{minipage}
\end{figure}

\begin{figure}[h]
 \hspace*{-0.37cm} % Negative space to shift left
    \makebox[0.95\textwidth][r]{ % Move content to the right edge of the line
        \begin{minipage}[t]{0.13\linewidth}
            \caption*{FourRooms}
            \includegraphics[width=\linewidth]{FourRooms_env.png} % 
        \end{minipage}
    }%
    
    % Bottom: Full-width heatmap
    \begin{minipage}[t]{\linewidth}
        \centering
        \includegraphics[width=\linewidth]{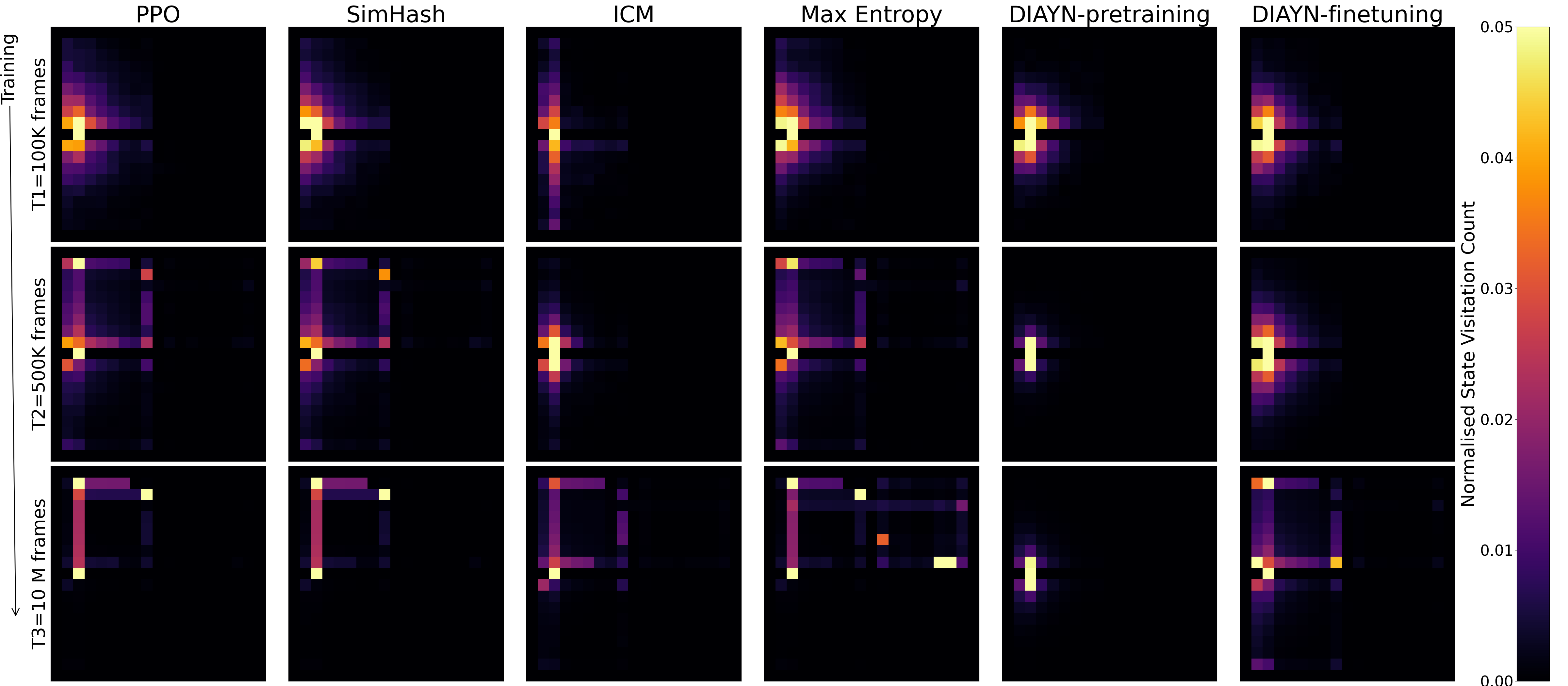} % Heatmap image file
        \caption{State visitation count during training for 10M frames on singleton FourRooms environment with RGB observations. For each intrinsic reward method, snapshots of the heatmap are taken at three different timesteps T1: 100K frames, T2: 500K frames and T3: 10M frames. Colour intensity represents the proportion of frames spent in each state, with high values capped for better visualisation.}
        \label{FouRooms_RGB}
    \end{minipage}
\end{figure}

\begin{figure}[h]
 \hspace*{-0.3cm} % Negative space to shift left
    \makebox[0.95\textwidth][r]{ % Move content to the right edge of the line
        \begin{minipage}[t]{0.15\linewidth}
            \caption*{RedBlueDoors}
            \includegraphics[width=\linewidth]{RedBlue_env.png} % 
        \end{minipage}
    }%
    
    % Bottom: Full-width heatmap
    \begin{minipage}[t]{\linewidth}
        \centering
        \includegraphics[width=\linewidth]{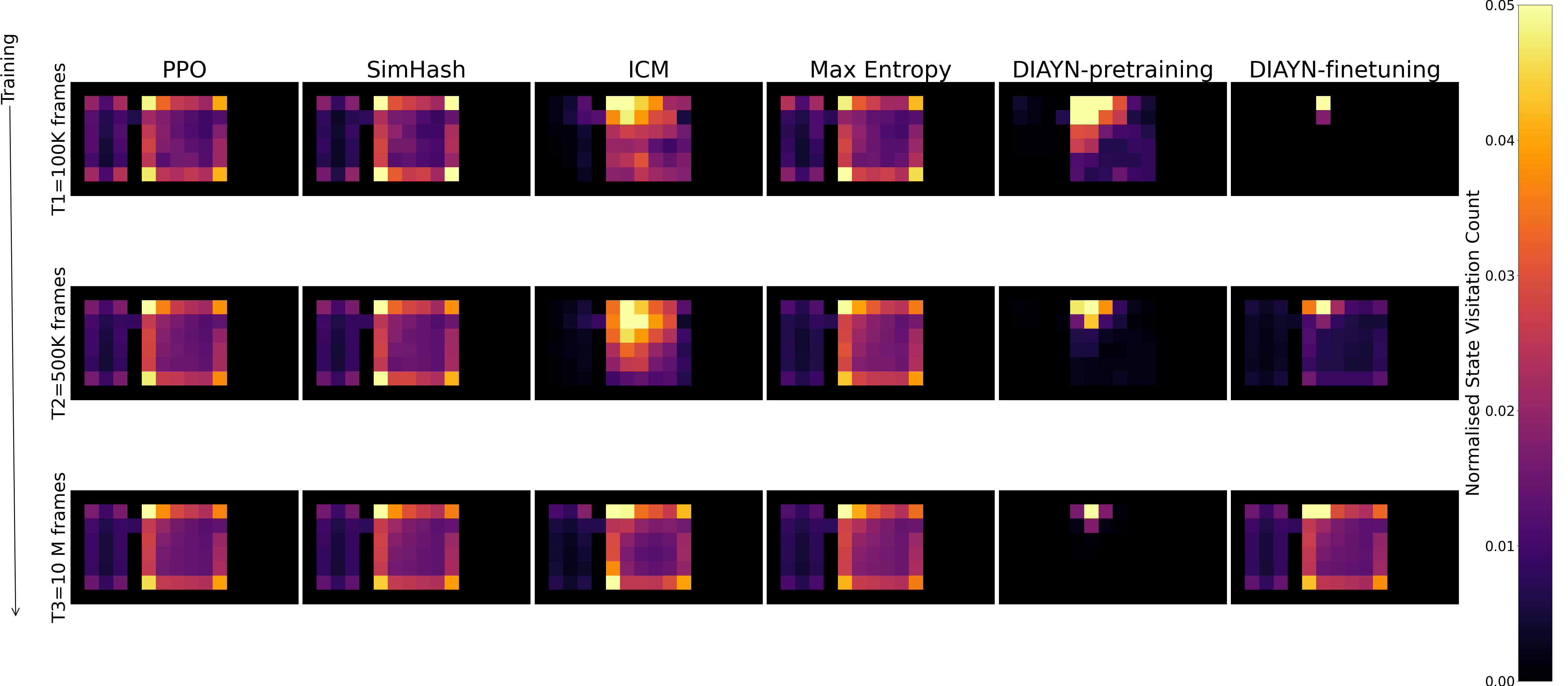} % Heatmap image file
        \caption{State visitation count during training for 10M frames on singleton RedBlueDoors environment with RGB observations. For each intrinsic reward method, snapshots of the heatmap are taken at three different timesteps T1: 100K frames, T2: 500K frames and T3: 10M frames. Colour intensity represents the proportion of frames spent in each state, with high values capped for better visualisation.}
        \label{RedBlueDoors_RGB}
    \end{minipage}
\end{figure}

\FloatBarrier
\section{DIAYN Extrinsic}
\begin{figure}[h]
\centering
\includegraphics[width=\textwidth]{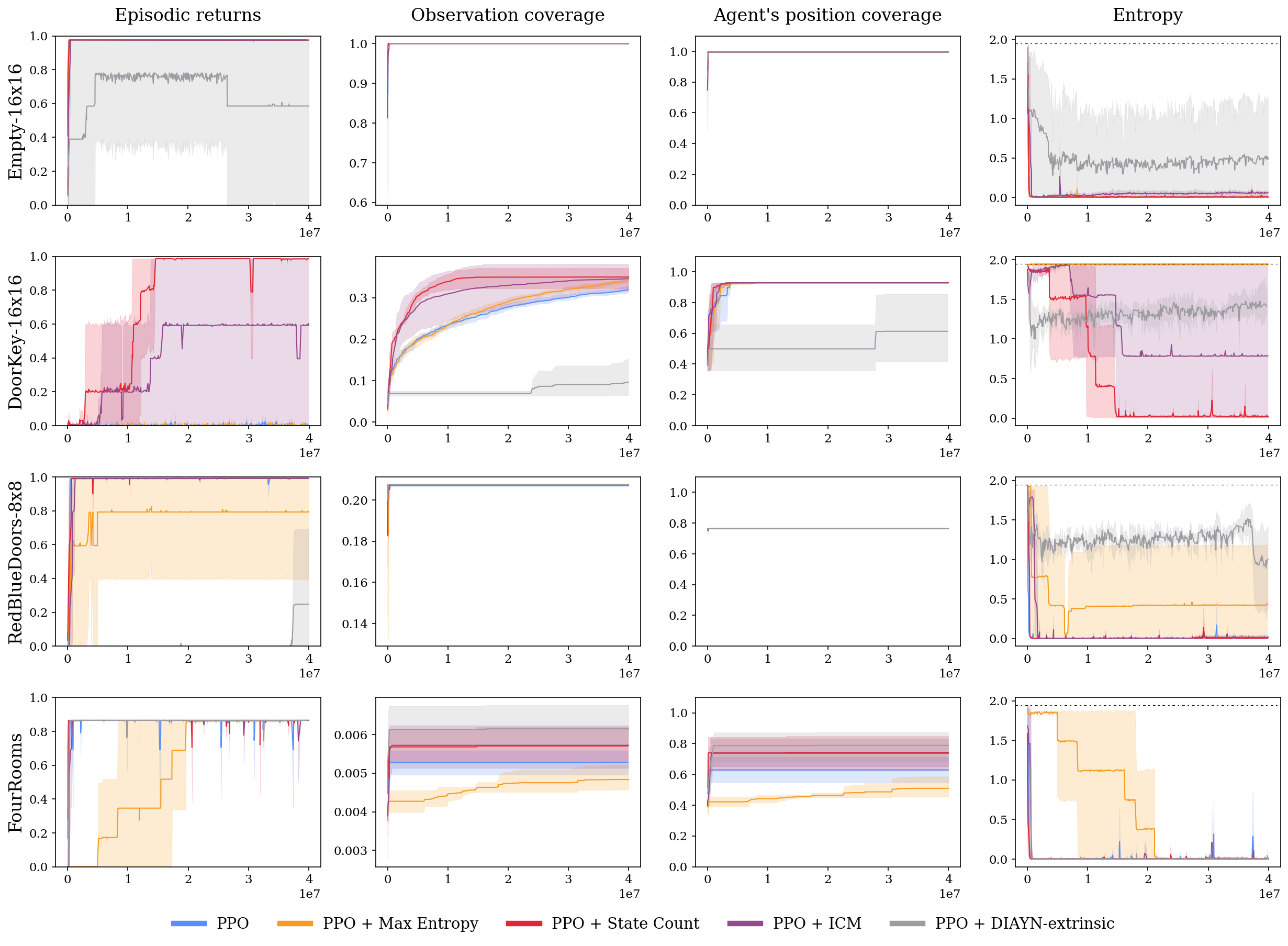}
\caption{Results for DIAYN combined with extrinsic rewards on Grid Encodings observation space.
}
\label{DIAYN+extrinsic}
\end{figure}

Initially, we evaluated DIAYN combined with extrinsic rewards, but it did not perform well because of the imbalance between discriminability and reward maximisation (see Figure \ref{DIAYN+extrinsic}). Recognising that DIAYN is primarily intended for unsupervised pretraining of skills rather than simultaneous use with return maximisation, we decided to split the training budget between pretraining and finetuning.
\label{DIAYN_extrinsic}

%%%%%%%%%%%%%%%%%%%%%%%%%%%%%%%%%%%%%%%%%%%%%%%

%\begin{figure}[h]
    % Top right: Empty environment image
%    \makebox[0.95\textwidth][r]{ % Move content to the right edge of the line
%        \begin{minipage}[t]{0.13\linewidth}
%            \caption*{Empty}
%            \includegraphics[width=\linewidth]{Empty_env.png} % Empty environment image file
%            \label{Empty}
%        \end{minipage}
%    }%

%    % Bottom: Full-width heatmap
%    \begin{minipage}[t]{\linewidth}
%        \centering
%        \includegraphics[width=\linewidth]{Empty_NonRGB_edited.png} % Heatmap image file
%        \caption{State visitation count}%~\cite{yarats2021reinforcement,nguyen2021sample,choi2021variational} }
%        \label{heatmap_Empty_NonRGB}
%    \end{minipage}
%\end{figure}
%\newpage

%%%%%%%%%%%%%%%%%%%%%%%%%%%%%%%%%%%%%%%%%%%%%%%%%%%%%%%%%%%%%%%%
%% Bibliography
%%%%%%%%%%%%%%%%%%%%%%%%%%%%%%%%%%%%%%%%%%%%%%%%%%%%%%%%%%%%%%%%
\bibliography{main}

\end{document}